\documentclass[runningheads]{llncs}

\usepackage{accv}

\usepackage{accvabbrv}

\usepackage{graphicx}
\usepackage{booktabs}

\usepackage[accsupp]{axessibility}  

\usepackage{hyperref}

\usepackage{orcidlink}

\usepackage{microtype}

\renewcommand{\paragraph}[1]{\vspace{.5em}\noindent\textbf{#1.}}

\usepackage[accsupp]{axessibility}
\usepackage{bbm}
\usepackage{amssymb,mathtools}
\usepackage{multirow}
\usepackage{colortbl}
\usepackage{soul}
\usepackage{comment}
\usepackage{makecell}
\usepackage{pifont}
\usepackage{caption}
\usepackage{wrapfig}
\usepackage{subcaption}
\usepackage{arydshln}

\newcommand{\cmark}{\ding{51}}%
\newcommand{\xmark}{\ding{55}}%

\usepackage[most]{tcolorbox}
\tcbuselibrary{listings, breakable, skins}

\definecolor{baselinecolor}{gray}{.92}
\newcommand{\baseline}[1]{\cellcolor{baselinecolor}{#1}}
\begin{document}

\title{DCR-TSF: Dual-Level Compositional Reasoning with Triadic Spherical Fusion for Training-Free Zero-Shot Composed Image Retrieval}

\titlerunning{DCR-TSF}

\author{Gwang-Ho Na\inst{1}\orcidlink{0009-0005-8543-8347} \and
Ho-Joong Kim\inst{1}\orcidlink{0000-0003-4200-5136} \and
Seong-Whan Lee\inst{1}\orcidlink{0000-0002-6249-4996}}

\authorrunning{G.-H. Na et al.}

\institute{
Dept. of Artificial Intelligence, Korea University, Seoul, Korea\\
\email{\{g\_h\_na, hojoong\_kim, sw.lee\}@korea.ac.kr}}


\maketitle

\newcommand{\modelname}{DCR-TSF}

\begin{abstract}
Zero-shot composed image retrieval~(ZS-CIR) aims to retrieve a target image that reflects an intended textual modification while maintaining semantic consistency with a reference image.
Existing training-free ZS-CIR methods typically rely on fine-grained generated target descriptions, which often overemphasize reference-specific visual details and lead to appearance-biased retrieval.
Moreover, since these descriptions are generated through multimodal large language models (MLLMs), the direct influence of the original image-text query can be weakened in the final retrieval representation.
In this paper, we propose DCR-TSF, a training-free ZS-CIR framework that combines Dual-level Compositional Reasoning (DCR) and Triadic Spherical Fusion (TSF).
DCR generates fine- and coarse-grained descriptions from the reference image and modification text, capturing complementary target-side semantics across different granularities.
To preserve the original image-text query information, TSF constructs a composed-query anchor by spherically interpolating the reference image feature and modification text feature, and fuses it with the fine- and coarse-grained representations using a weighted Karcher mean on the unit hypersphere.
The resulting final query representation balances detailed visual evidence, high-level semantic abstraction, and direct preservation of the original composed query.
Extensive experiments on CIRCO, CIRR, and Fashion-IQ demonstrate that our method achieves state-of-the-art performance among training-free methods.
\end{abstract}
\section{Introduction}
\label{sec:introduction}
\begin{figure}[!t]
    \centering
    \begin{subfigure}[t]{0.49\linewidth}
        \centering
        \includegraphics[width=\linewidth]{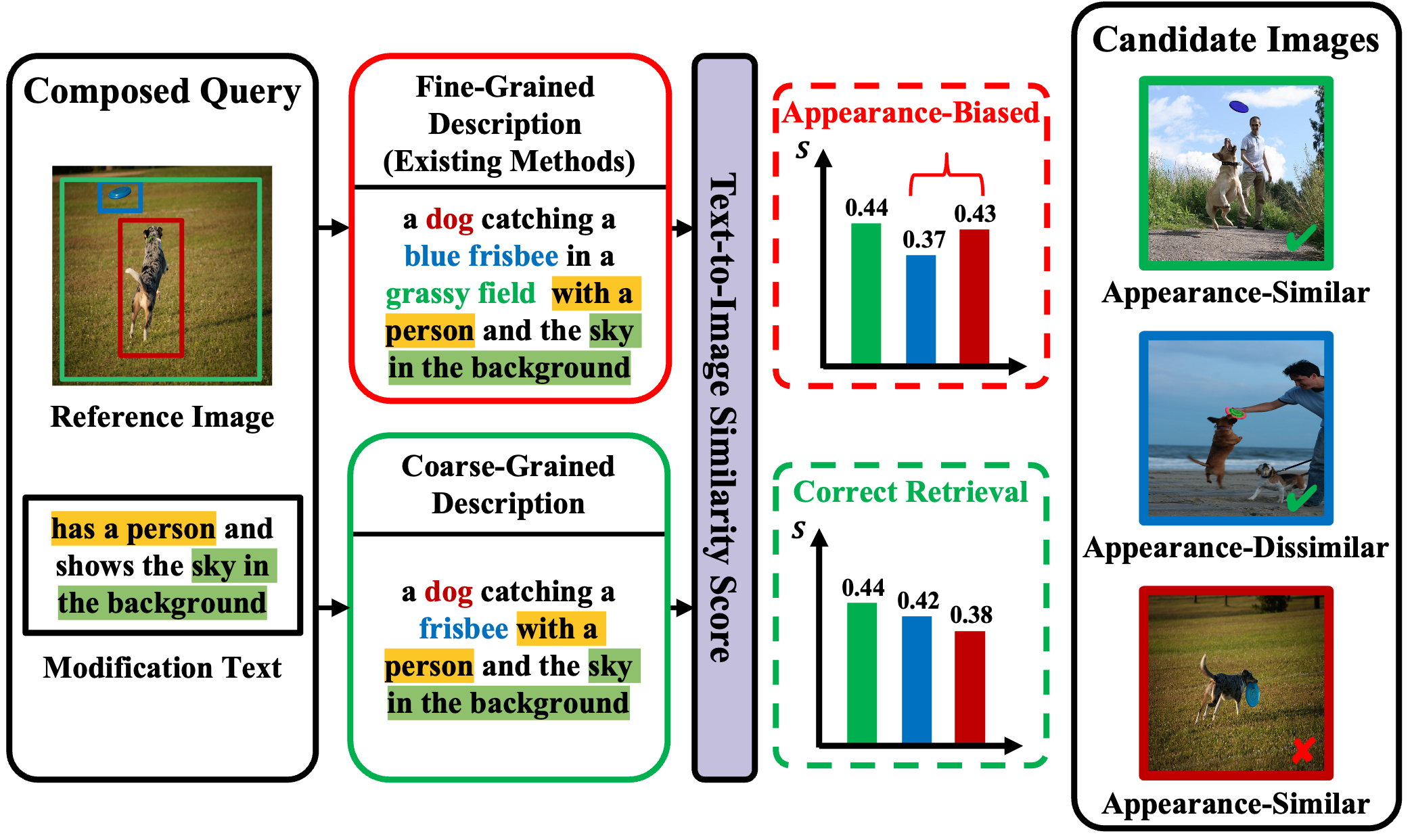}
        \caption{Appearance-biased case}
        \label{fig:motivation_a}
    \end{subfigure}%
    \hfill
    \begin{subfigure}[t]{0.49\linewidth}
        \centering
        \includegraphics[width=\linewidth]{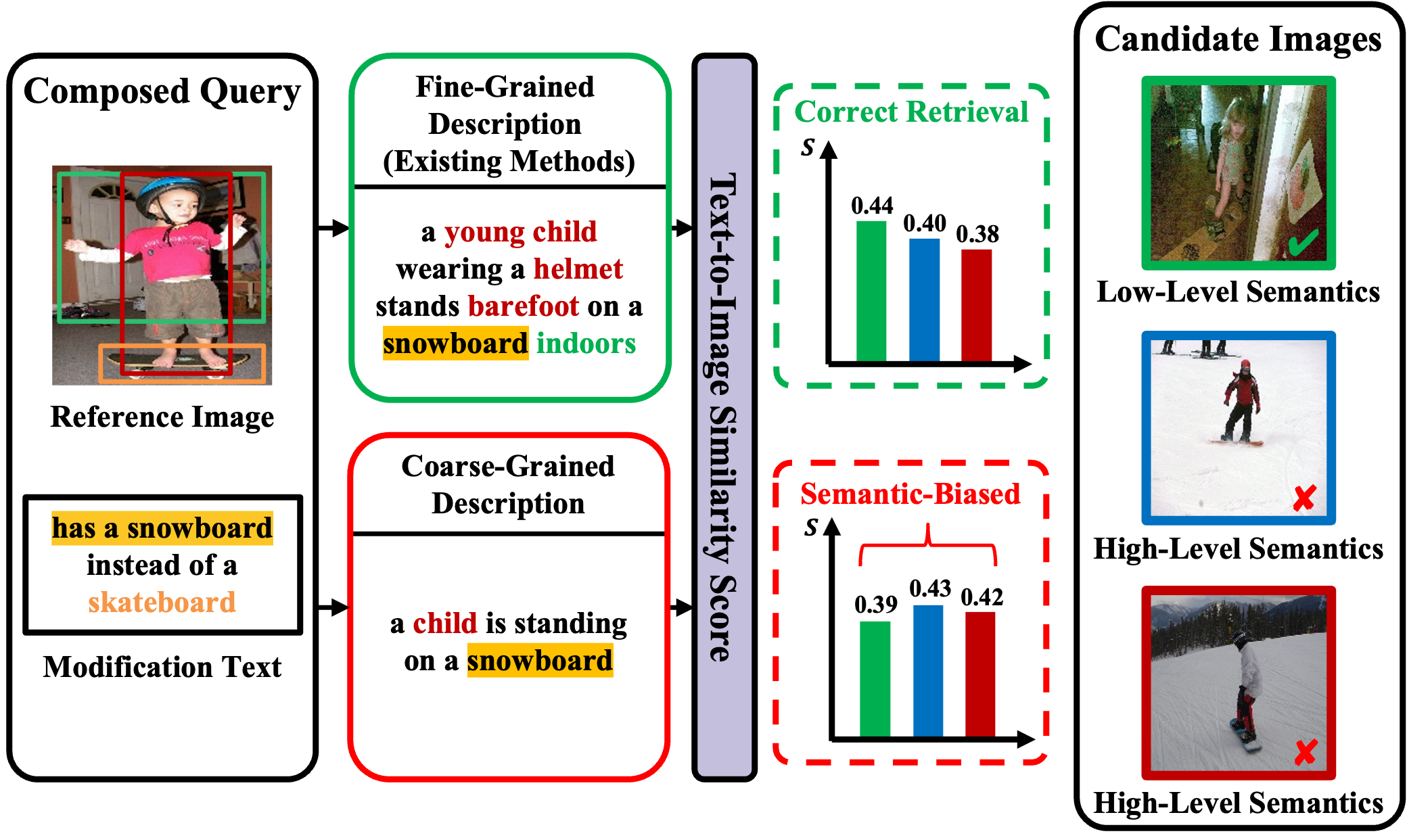}
        \caption{Information deficient case}
        \label{fig:motivation_b}
    \end{subfigure}%
    \vspace{-1pt}
    \caption{\textbf{Biases in Fine- and Coarse-grained Descriptions for ZS-CIR.} \textcolor{gray}{Colored text} refers to observed elements in the reference image. \colorbox{baselinecolor}{Background color} indicates phrases appearing in the modification text.
    The bar colors in the text-to-image similarity score plots correspond to the candidate images.
    (a): Fine-grained description is biased towards the visual details of the reference image. (b): Coarse-grained description lacks details of the reference image.}
    \label{fig:motivation}
\end{figure}

Zero-shot composed image retrieval~(ZS-CIR)~\cite{CIRR,PALAVRA,Pic2Word,CIRCO} is the task of retrieving an image that reflects a user-specified textual modification of a given reference image while maintaining semantic consistency.
Unlike conventional image retrieval~\cite{4_pill, 5_lim2000text}, ZS-CIR operates under a fuzzy matching paradigm, which means multiple different images can match with the modification text provided by the user.
This inherent ambiguity makes it difficult to construct large-scale annotated datasets for ZS-CIR due to the combinatorial explosion of possible reference-modification pairs.
To address this issue, recent works~\cite{LDRE,SEIZE,CoTMR} have explored training-free paradigms leveraging pretrained vision-language models.
Among them, existing ZS-CIR methods~\cite{LDRE,SEIZE,CoTMR} commonly adopt a text-guided matching approach that generates a target image description predicting what the target image should look like using multimodal large language models (MLLMs)~\cite{Qwen2.5,InternVL3.5}.
However, existing methods still suffer from a fundamental limitation: they rely on appearance-biased fine-grained descriptions that overemphasize reference-specific visual details.

To demonstrate the issue of existing fine-grained descriptions, we additionally construct a coarse-grained description that describes only the key modification-related elements while omitting reference-specific details.
Fig.~\ref{fig:motivation} shows how different granularities of generated target-image descriptions affect retrieval results.
As shown in Fig.~\ref{fig:motivation}(a), the existing fine-grained description overemphasizes reference-specific appearance details such as the \emph{blue frisbee} and \emph{grassy field}.
This overemphasis introduces a visual bias, causing CLIP~\cite{CLIP} to highlight visually similar but semantically irrelevant candidates (e.g., the third image).
In contrast, the coarse-grained description excludes these reference-specific details such as \emph{blue} and \emph{grassy field}, retrieving correct images.
Conversely, Fig.~\ref{fig:motivation}(b) shows the complementary information-deficient case for coarse-grained description.
The coarse-grained description only includes \emph{child} and \emph{snowboard}.
In this case, the modification text only specifies ``has a snowboard instead of a skateboard,'' where visual cues in the reference image such as \emph{young}, \emph{helmet}, \emph{barefoot}, and \emph{indoors} play a critical role in understanding the modification intent.
Since the coarse-grained description omits these details, the coarse-grained case fails to capture the fine-grained semantics.
These examples illustrate that relying on a single reasoning granularity inevitably introduces two complementary failure cases.

In this paper, we propose Dual-level Compositional Reasoning with Triadic Spherical Fusion (\modelname) for training-free ZS-CIR.
Our method captures dual-granular textual reasoning implied by the reference image and modification text, addressing the complementary limitations of appearance-biased fine-grained descriptions and information-deficient coarse-grained descriptions.
To achieve this, \modelname\ first generates fine- and coarse-grained textual representations from the reference image and modification text.
The fine-grained representation preserves detailed appearance cues, while the coarse-grained representation abstracts semantic intent at a conceptual level.
However, because the fine- and coarse-grained representations are obtained through MLLM-generated descriptions, the direct influence of the original image-text query can be weakened in the final retrieval representation.
To explicitly preserve the original composed query, we introduce Triadic Spherical Fusion (TSF).
TSF first constructs a composed-query anchor by spherically interpolating the reference image feature and the modification text feature.
TSF then fuses the fine-grained representation, coarse-grained representation, and composed-query anchor using a weighted Karcher mean~\cite{Karchermean} on the unit hypersphere.
The resulting final query representation balances visual detail, semantic abstraction, and direct preservation of the original image-text query.
Extensive comparisons and ablation studies on Fashion-IQ~\cite{FashionIQ}, CIRR~\cite{CIRR}, and CIRCO~\cite{CIRCO} demonstrate the effectiveness of \modelname, which achieves state-of-the-art performance among training-free ZS-CIR methods.

\noindent Our contributions are summarized as three-fold:
\begin{itemize}
    \item We introduce Dual-level Compositional Reasoning (DCR), which generates fine- and coarse-grained textual representations to preserve detailed visual attributes and capture high-level semantic intent.

    \item We propose Triadic Spherical Fusion (TSF), which preserves the original composed-query information by fusing a composed-query anchor with fine- and coarse-grained representations on the unit hypersphere.

    \item Extensive comparisons and ablation studies on three standard benchmarks demonstrate the effectiveness of \modelname, which achieves state-of-the-art performance among training-free ZS-CIR methods.
\end{itemize}

\section{Related Work}
\label{sec:related_work}

\noindent \textbf{Composed Image Retrieval (CIR).}
Early supervised CIR methods~\cite{CIRR,ARTEMIS} jointly learned multimodal embedding spaces from annotated reference-target pairs, requiring large-scale labeled CIR datasets.
Due to the difficulty of collecting triplet labels, recent studies have explored ZS-CIR, which performs retrieval without task-specific training.
A basic approach in ZS-CIR is to construct a composed query representation by combining the reference image feature and the modification text feature in the CLIP~\cite{CLIP} embedding space.
This composition is commonly implemented by weighted feature aggregation, while spherical linear interpolation can also be used to better respect the unit-hypersphere geometry of normalized CLIP embeddings~\cite{TAT}.
With the progress of MLLMs~\cite{InternVL3.5,Qwen2.5}, another line of training-free ZS-CIR methods has emerged, which generates target descriptions by jointly interpreting the reference image and modification text.
CIReVL~\cite{CIReVL} adopts a two-stage pipeline that uses a captioner and a large language model to generate a target description.
LDRE~\cite{LDRE} and SEIZE~\cite{SEIZE} improve description-based retrieval by leveraging multiple generated descriptions.
OSrCIR~\cite{OSrCIR} introduces a chain-of-thought one-stage framework to mitigate the potential information loss of a two-stage pipeline.
IP-CIR~\cite{IP-CIR} leverages a diffusion model to generate a proxy image, directly capturing visual changes.
CoTMR~\cite{CoTMR} utilizes an MLLM to infer existent and nonexistent objects and incorporates these reasoning results into the retrieval process.
Despite their effectiveness, existing description-based ZS-CIR methods generally rely on fine-grained target descriptions or multiple generated descriptions centered on fine-grained visual details.
This reliance can overemphasize reference-specific appearance cues, leading to appearance-biased retrieval.
Furthermore, because these retrieval representations are derived from MLLM-generated descriptions, the direct influence of the original image-text query can be weakened in the final retrieval representation.
In contrast, our method combines Dual-level Compositional Reasoning (DCR) with Triadic Spherical Fusion (TSF), using coarse-grained descriptions to complement fine-grained target semantics while explicitly preserving the original composed query through a composed-query anchor.

\noindent \textbf{Vision-Language Models (VLMs).}
VLMs demonstrate strong cross-modal reasoning capabilities, enabling zero-shot generalization across a wide range of visual-linguistic tasks.
CLIP~\cite{CLIP} pioneers large-scale contrastive pretraining on image-text pairs, establishing a unified embedding space that effectively bridges visual and textual information.
Recent MLLMs, such as InternVL~\cite{InternVL3.5} and Qwen-VL~\cite{Qwen2.5}, further advance fine-grained perception and instruction following, enabling contextual reasoning over image content.
These advances make it possible to generate complementary textual representations without any task-specific supervision.
Our method leverages pretrained VLMs for image-text embedding and MLLMs for generating fine- and coarse-grained descriptions.
\section{Method}
\label{sec:method}

\subsection{Preliminary}
\label{subsec:preliminary}
Let a composed query be defined as $Q = \{I_r, D_m\}$, where $I_r$ denotes the reference image and $D_m$ denotes the modification text.
Given $N_D$ candidate images $\{I^{(i)}_t\}_{i=1}^{N_D}$, the goal of CIR is to retrieve the top-$K$ target images that best match the query, with $K \ll N_D$.
To enable cross-modal comparison between the query and candidate images, all visual and textual inputs are projected into a shared multimodal embedding space using CLIP~\cite{CLIP}.

\begin{figure*}[!th]
    \centering
    \includegraphics[width=\linewidth]{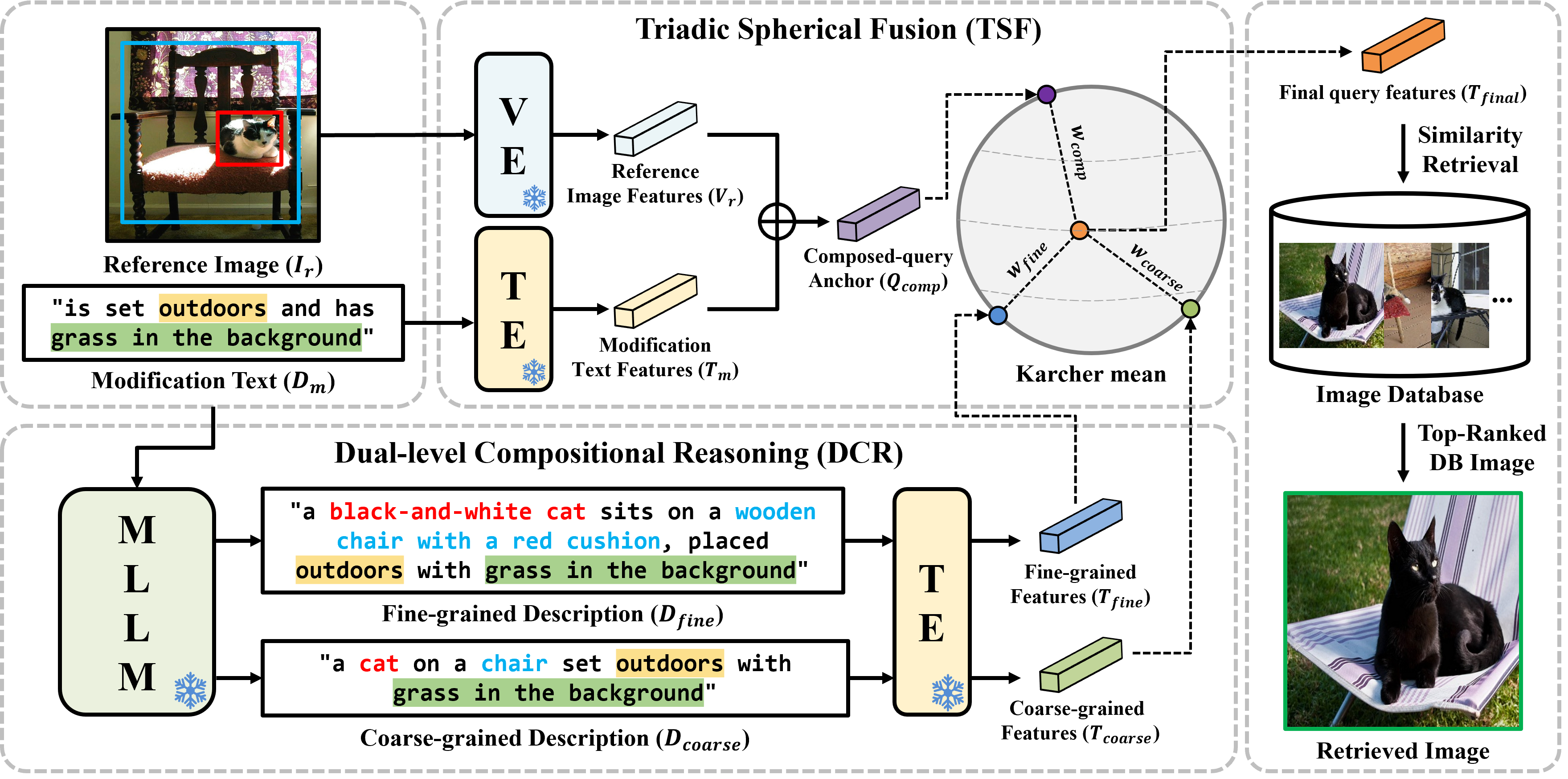}
    \caption{\textbf{Overview of \modelname.}
    Given a reference image $I_r$ and a modification text $D_m$, DCR generates fine- and coarse-grained descriptions, which are encoded into $T_{\text{fine}}$ and $T_{\text{coarse}}$.
    TSF constructs a composed-query anchor $Q_{\text{comp}}$ from $V_r$ and $T_m$, and fuses it with $T_{\text{fine}}$ and $T_{\text{coarse}}$ using a weighted Karcher mean~\cite{Karchermean}.
    The resulting final query representation $T_{\text{final}}$ is used to compute the final retrieval score $S^{(i)}$.}
    \label{fig:method}
\end{figure*}

\subsection{\modelname}
In this part, we briefly describe \modelname, which mainly consists of two parts:
(1) Dual-level Compositional Reasoning (DCR), which generates fine- and coarse-grained descriptions to capture complementary target-side semantics;
(2) Triadic Spherical Fusion (TSF), which constructs a composed-query anchor and fuses it with the fine- and coarse-grained representations on the unit hypersphere. The overall architecture of \modelname\ is illustrated in Fig.~\ref{fig:method}.

\subsection{Dual-level Compositional Reasoning (DCR)}
\begin{figure}[!t]
    \centering
    \includegraphics[width=1.0\linewidth]{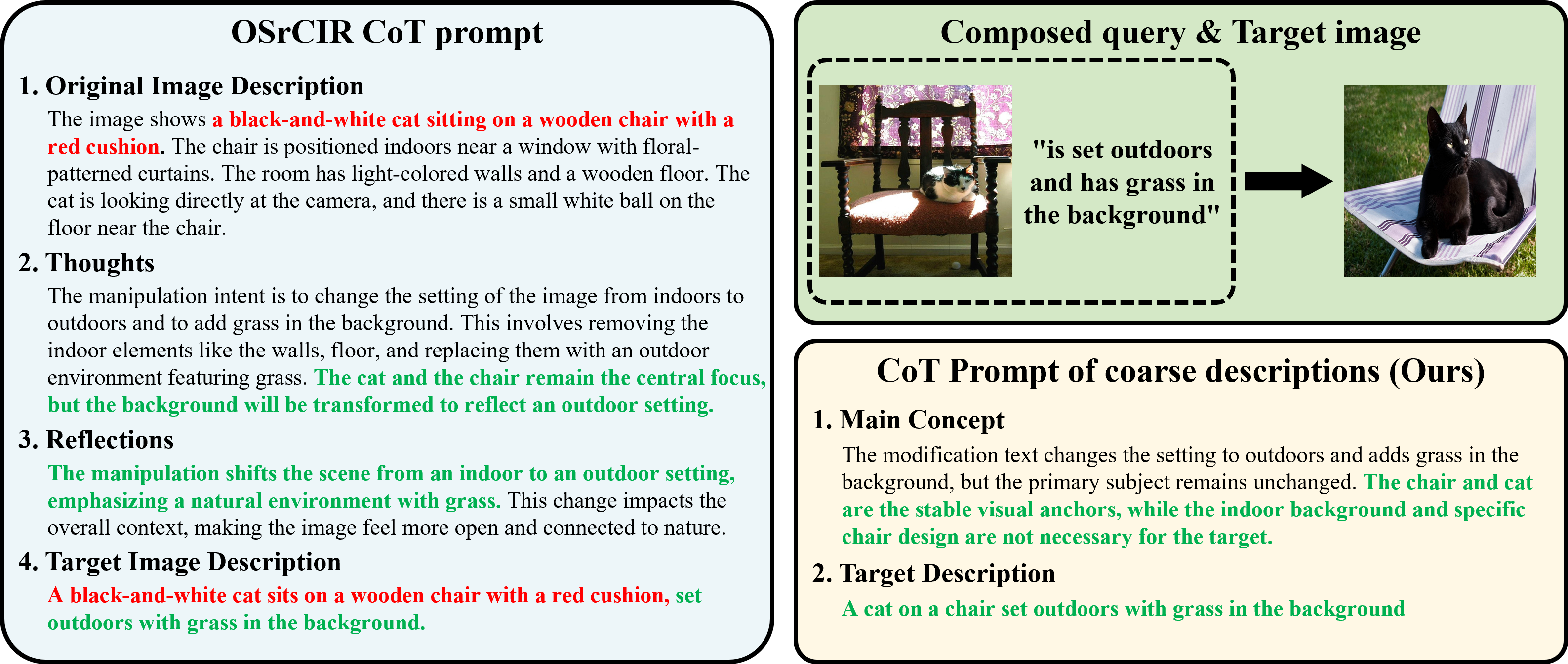}
    \caption{\textbf{Comparison between OSrCIR and our coarse-grained description prompt.}
    OSrCIR~\cite{OSrCIR} generates a detailed target description through a CoT prompt, preserving rich reference-specific visual evidence that may be irrelevant to the modification condition.
    In contrast, our coarse-grained description prompt identifies stable visual anchors and focuses on modification-related semantics, omitting excessive reference-specific details and providing a complementary abstraction for DCR.}
    \label{fig:prompt_comparison}
\end{figure}

\noindent \textbf{Dual-level Description Generation.}
Following previous description-based methods~\cite{OSrCIR,CoTMR,SEIZE}, 
we exploit an MLLM to infer target-image descriptions from the reference image $I_r$ and the modification text $D_m$.
The modification text $D_m$ represents a user-specified textual instruction.
We first generate a fine-grained description $D_{\text{fine}}$, following the conventional target-description generation scheme used in existing ZS-CIR methods.
This description preserves detailed visual evidence from the reference image and describes what the target image should look like.
However, as discussed in Sec.~\ref{sec:introduction}, fine-grained descriptions can overemphasize reference-specific visual details that are not directly relevant to the intended modification.
To complement this, we additionally generate a coarse-grained description $D_{\text{coarse}}$, which describes the target image at a more abstract semantic level by focusing on stable visual anchors and key modification-related elements.
Fig.~\ref{fig:prompt_comparison} compares our coarse-grained description generation with OSrCIR~\cite{OSrCIR}.
It shows that our prompt suppresses excessive reference-specific details and produces a coarse-grained description complementary to the fine-grained description.

The two descriptions are generated as follows:
\begin{equation}
    D_{\text{coarse}} = \text{MLLM}(I_r, D_m, P_{\text{coarse}}), \quad
    D_{\text{fine}} = \text{MLLM}(I_r, D_m, P_{\text{fine}}),
\end{equation}
where $P_{\text{coarse}}$ and $P_{\text{fine}}$ denote the prompting templates for coarse- and fine-grained generation, respectively.
$D_{\text{coarse}}$ captures global, scene-level semantics and the intended modification, while $D_{\text{fine}}$ preserves local visual details in the reference image.
Detailed prompts for $P_\text{coarse}$ and $P_\text{fine}$ are provided in the supplementary material.

\noindent \textbf{Dual-level Representation Encoding.}
The coarse-grained description $D_\text{coarse}$ and fine-grained description $D_\text{fine}$ are fed into the frozen text encoder of CLIP~\cite{CLIP} to obtain their corresponding representations:
\begin{equation}
T_{\text{coarse}} = \mathrm{TE}(D_{\text{coarse}})\text{,} \quad
T_{\text{fine}} = \mathrm{TE}(D_{\text{fine}})\text{,}
\end{equation}
where $\mathrm{TE}(\cdot)$ denotes the CLIP~\cite{CLIP} text encoder.
$T_{\text{coarse}}$ captures global, concept-level semantics and the intended modification, while $T_{\text{fine}}$ preserves local visual details grounded in the reference image.

\subsection{Triadic Spherical Fusion (TSF)}

\noindent \textbf{Composed-query Anchor.}
Although $T_{\text{fine}}$ and $T_{\text{coarse}}$ provide complementary target-side semantics, they are derived from MLLM-generated descriptions.
Therefore, relying only on these representations may dilute the original image-text query information.
To explicitly preserve the original composed query, we construct a composed-query anchor from the reference image and modification text.
Specifically, we first extract the reference image feature $V_r$ and the modification text feature $T_m$:
\begin{equation}
V_r = \mathrm{VE}(I_r)\text{,} \quad
T_m = \mathrm{TE}(D_m)\text{,}
\end{equation}
where $\mathrm{VE}(\cdot)$ and $\mathrm{TE}(\cdot)$ denote the CLIP visual and text encoders, respectively.
Since CLIP features are normalized on the unit hypersphere, we use Slerp~\cite{SLERP} to interpolate the reference image feature and the modification text feature:
\begin{equation}
Q_{\text{comp}} =
\frac{\sin((1-\lambda)\theta)}{\sin(\theta)}V_r
+
\frac{\sin(\lambda\theta)}{\sin(\theta)}T_m\text{,}
\end{equation}
where $\theta = \cos^{-1}(V_r \cdot T_m)$ is the angle between the reference image feature and the modification text feature, and $\lambda$ is a hyperparameter that balances the contribution of $V_r$ and $T_m$.
A larger $\lambda$ emphasizes the modification text feature $T_m$, while a smaller $\lambda$ preserves more reference-level visual information from $V_r$.
The resulting $Q_{\text{comp}}$ serves as a composed-query anchor that directly preserves the original image-text query information.

\noindent \textbf{Triadic Spherical Fusion.}
To obtain the final query representation, TSF fuses three complementary representations: the fine-grained representation $T_{\text{fine}}$, the coarse-grained representation $T_{\text{coarse}}$, and the composed-query anchor $Q_{\text{comp}}$.
Since sequentially applying Slerp to multiple representations can introduce order dependence, we use a weighted Karcher mean~\cite{Karchermean} on the unit hypersphere.
Unlike Euclidean averaging, the Karcher mean finds a point on the unit hypersphere that minimizes the weighted sum of squared geodesic distances to the three representations:
\begin{equation}
\begin{aligned}
T_{\text{final}}
=
\arg\min_{\|Y\|_2=1}
&\;
w_{\text{fine}}
\left(\cos^{-1}(Y^\top T_{\text{fine}})\right)^2
+
w_{\text{coarse}}
\left(\cos^{-1}(Y^\top T_{\text{coarse}})\right)^2  \\
&+
w_{\text{comp}}
\left(\cos^{-1}(Y^\top Q_{\text{comp}})\right)^2,
\end{aligned}
\end{equation}
where $Y$ is the unit-norm optimization variable, and $w_{\text{fine}}$, $w_{\text{coarse}}$, and $w_{\text{comp}}$ are fusion weights satisfying $w_{\text{fine}}+w_{\text{coarse}}+w_{\text{comp}}=1$.
Here, $\cos^{-1}(a^\top b)$ computes the geodesic distance between two unit-normalized representations on the hypersphere.
In practice, we compute this mean using an iterative solver based on logarithmic and exponential maps.
This fusion produces the final query representation $T_{\text{final}}$, which balances detailed visual evidence, high-level semantic abstraction, and direct preservation of the original composed query.
More details are provided in the supplementary material.


\noindent \textbf{Final Retrieval Score.}
Given each candidate image $I_t^{(i)}$, we extract its visual feature.
The final retrieval score is computed as the cosine similarity between the final query representation and each candidate visual feature:
\begin{equation}
V_t^{(i)} = \mathrm{VE}(I_t^{(i)})\text{,}
\qquad
S^{(i)} =
\frac{V_t^{(i)} \cdot T_{\text{final}}}
{\|V_t^{(i)}\|\,\|T_{\text{final}}\|}\text{.}
\end{equation}
Finally, all candidate images are ranked by the final retrieval score $S^{(i)}$.

\section{Experiments}
\label{sec:experiments}

\begin{table}[!t]
\centering
\caption{\textbf{Comparison with the state-of-the-art methods on the CIRCO and CIRR test sets.} Best results are highlighted in bold, and the second best are underlined.}
\label{table:main}
\resizebox{0.95\linewidth}{!}{
\begin{tabular}{llc cccccccc c cccc}

\Xhline{2\arrayrulewidth}
\noalign{\smallskip}

\multirow{2.7}{*}{\textbf{Backbone}} & \multirow{2.7}{*}{\textbf{Method}} & \multirow{2.7}{*}{\shortstack{\textbf{Training}\\ \textbf{-free}}} & \multicolumn{4}{c}{\textbf{CIRCO}} & & \multicolumn{7}{c}{\textbf{CIRR}} \\

\noalign{\smallskip}
\cline{4-7}
\cline{9-15}
\noalign{\smallskip}

& & & mAP@5 & mAP@10 & mAP@25 & mAP@50 & & R@1 & R@5 & R@10 & R\textsubscript{sub}@1 & R\textsubscript{sub}@2 & R\textsubscript{sub}@3 & Avg.\\

\noalign{\smallskip}
\Xhline{2\arrayrulewidth}
\noalign{\smallskip}

\multirow{9}{*}{ViT-B/32} 
 & PALAVRA~\cite{PALAVRA} & \xmark & 4.61 & 5.32 & 6.33 & 6.80 & & 16.62 & 43.49 & 58.51 & 41.61 & 65.30 & 80.94 & 42.55 \\
 & SEARLE~\cite{CIRCO} & \xmark & 9.35 & 9.94 & 11.13 & 11.84 & & 24.00 & 53.42 & 66.82 & 54.89 & 76.60 & 88.19 & 54.16\\
 & Slerp~\cite{TAT} & \cmark & 6.35 & 7.11 & 8.12 & 8.75 & & 24.22 & 50.94 & 63.49 & 57.86 & 78.27 & 89.25 & 54.40\\
 & CIReVL~\cite{CIReVL} & \cmark & 14.94 & 15.42 & 17.00 & 17.82 & & 23.94 & 52.51 & 66.00 & 60.17 & 80.05 & 90.19 & 56.34\\
 & LDRE~\cite{LDRE} & \cmark & 17.96 & 18.32 & 20.21 & 21.11 & & 25.69 & 55.13 & 69.04 & 60.53 & 80.65 & 90.70 & 57.83\\
 & OSrCIR~\cite{OSrCIR} & \cmark & 18.04 & 19.17 & 20.94 & 21.85 & & 25.42 & 54.54 & 68.19 & 62.31 & 80.86 & 91.13 & 58.43\\
 & SEIZE \cite{SEIZE} & \cmark & 19.04 & 19.64 & 21.55 & 22.49 & & 27.47 & 57.42 & 70.17 & 65.59 & 84.48 & \underline{92.77} & 61.51\\
 & CoTMR \cite{CoTMR} & \cmark & \underline{22.23} & \underline{22.78} & \underline{24.68} & \underline{25.74} & & \underline{31.50} & \underline{60.80} & \underline{73.04} & \underline{66.61} & \underline{84.50} & 92.55 & \underline{63.71}\\
 & \baseline{\textbf{\modelname\ (Ours)}} & \baseline{\cmark} & \baseline{\textbf{25.07}} & \baseline{\textbf{25.81}} & \baseline{\textbf{28.05}} & \baseline{\textbf{29.09}} & \baseline{} & \baseline{\textbf{35.93}} & \baseline{\textbf{67.28}} & \baseline{\textbf{79.28}} & \baseline{\textbf{67.66}} & \baseline{\textbf{84.99}} & \baseline{\textbf{93.74}} & \baseline{\textbf{67.47}}\\
 
\noalign{\smallskip}
\hline
\noalign{\smallskip}

\multirow{9}{*}{ViT-L/14}
 & SEARLE \cite{CIRCO} & \xmark & 11.68 & 12.73 & 14.33 & 15.12 & & 24.24 & 52.48 & 66.29 & 53.76 & 75.01 & 88.19 & 53.12 \\
 & Slerp~\cite{TAT} & \cmark & 8.76 & 9.84 & 11.30 & 11.99 & & 24.43 & 49.93 & 62.29 & 57.71 & 77.59 & 88.80 & 53.82\\
 & CIReVL \cite{CIReVL} & \cmark & 18.57 & 19.01 & 20.89 & 21.80 & & 24.55 & 52.31 & 64.92 & 59.54 & 79.88 & 89.69 & 55.92\\
 & LDRE \cite{LDRE} & \cmark & 23.35 & 24.03 & 26.44 & 27.50 & & 26.53 & 55.57 & 67.54 & 60.43 & 80.31 & 89.90 & 58.00 \\
 & OSrCIR~\cite{OSrCIR} & \cmark & 23.87 & 25.33 & 27.84 & 28.97 & & 29.45 & 57.68 & 69.86 & 62.12 & 81.92 & 91.10 & 59.90\\
 & SEIZE~\cite{SEIZE} & \cmark & 24.98 & 25.82 & 28.24 & 29.35 & & 28.65 & 57.16 & 69.23 & 66.22 & 84.05 & 92.34 & 61.69\\
 & IP-CIR~\cite{IP-CIR} & \cmark & 26.43 & 27.41 & 29.87 & 31.07 & & 29.76 & 58.82 & 71.21 & 62.48 & 81.64 & 90.89 & 60.65\\
 & CoTMR~\cite{CoTMR} & \cmark & \underline{27.61} & \underline{28.22} & \underline{30.61} & \underline{31.70} & & \underline{35.02} & \underline{64.75} & \underline{76.18} & \underline{69.39} & \underline{85.75} & \underline{93.33} & \underline{67.07}\\
 & \baseline{\textbf{\modelname\ (Ours)}} & \baseline{\cmark} & \baseline{\textbf{31.80}} & \baseline{\textbf{32.65}} & \baseline{\textbf{35.15}} & \baseline{\textbf{36.25}} & \baseline{} & \baseline{\textbf{39.13}} & \baseline{\textbf{69.95}} & \baseline{\textbf{80.94}} & \baseline{\textbf{70.34}} & \baseline{\textbf{86.53}} & \baseline{\textbf{93.81}} & \baseline{\textbf{70.15}}\\
 
\noalign{\smallskip}
\hline
\noalign{\smallskip}

\multirow{7}{*}{ViT-G/14} & CIReVL \cite{CIReVL} & \cmark & 26.77 & 27.59 & 29.96 & 31.03 & & 34.65 & 64.29 & 75.06 & 67.95 & 84.87 & 93.21 & 66.12\\
 & LDRE \cite{LDRE} & \cmark & 31.12 & 32.24 & 34.95 & 36.03 & & 36.15 & 66.39 & 77.25  & 68.82 & 85.66 & 93.76 & 67.60\\
 & OSrCIR~\cite{OSrCIR} & \cmark & 30.47 & 31.14 & 35.03 & 36.59 & & 37.26 & 67.25 & 77.33 & 69.22 & 85.28 & 93.55 & 68.24\\
 & SEIZE \cite{SEIZE} & \cmark & 32.46 & 33.77 & 36.46 & 37.55 & & 38.87 & 69.42 & 79.42 & \textbf{74.15} & \textbf{89.23} & \textbf{95.71} & \textbf{71.79}\\
 & IP-CIR~\cite{IP-CIR} & \cmark & \underline{32.75} & \underline{34.26} & \underline{36.86} & \underline{38.03} & & \underline{39.25} & \underline{70.07} & \underline{80.00} & 69.96 & 86.87 & 94.22 & 70.02\\
 & CoTMR \cite{CoTMR} & \cmark & 32.23 & 32.72 & 35.60 & 36.83 & & 36.36 & 67.52 & 77.82 & \underline{71.19} & 86.34 & 93.87 & 69.36\\
 & \baseline{\textbf{\modelname\ (Ours)}} & \baseline{\cmark} & \baseline{\textbf{37.47}} & \baseline{\textbf{38.63}} & \baseline{\textbf{41.26}} & \baseline{\textbf{42.45}} & \baseline{} & \baseline{\textbf{41.40}} & \baseline{\textbf{71.40}} & \baseline{\textbf{81.59}} & \baseline{{70.27}} & \baseline{\underline{87.30}} & \baseline{\underline{94.36}} & \baseline{\underline{70.84}} \\

\noalign{\smallskip}
\Xhline{2\arrayrulewidth}
\noalign{\smallskip}
\end{tabular}
}
\end{table}

\subsection{Setup}
\label{subsec:setup}
\noindent \textbf{Datasets.}
We evaluate our method on three CIR benchmarks: CIRCO~\cite{CIRCO}, CIRR~\cite{CIRR}, and Fashion-IQ~\cite{FashionIQ}.
CIRCO is constructed from real-world images in the COCO 2017 unlabeled set~\cite{COCO} and is the first CIR dataset to provide multiple ground-truth targets per query.
CIRR is the first large-scale natural image dataset specifically curated for CIR. 
Fashion-IQ is a fashion-domain dataset consisting of garment images categorized into dresses, shirts, and tops/tees. 

\noindent \textbf{Evaluation Metrics.}
For Fashion-IQ~\cite{FashionIQ} and CIRR~\cite{CIRR}, we adopt Recall@K (R@K), which measures the fraction of queries where the ground-truth item appears among the top-$K$ retrieved results.
We also report R\textsubscript{mean}, defined as the average of all R@K values, representing overall retrieval performance.
For CIRCO~\cite{CIRCO}, we use mean Average Precision at $K$ (mAP@K).  
For CIRR~\cite{CIRR}, we report R@K, Recall\textsubscript{subset}@K (R\textsubscript{sub}@K), and the average score (Avg.) following~\cite{CIRR}.
R\textsubscript{sub}@K evaluates retrieval accuracy within a curated subset of visually similar candidates.
Following the previous works~\cite{CIRR,CoTMR}, the Avg. metric is defined as the mean of R@5 and R\textsubscript{sub}@1 for CIRR.


\noindent \textbf{Implementation Details.}
For the MLLM backbone, we employ Qwen2.5-VL-32B~\cite{Qwen2.5}.
For the image and text embeddings, following previous methods~\cite{OSrCIR,CoTMR}, we conduct experiments using OpenCLIP~\cite{OpenCLIP} models, including ViT-B/32, ViT-L/14, and ViT-G/14.
Our experiments are conducted on four NVIDIA A6000 GPUs.
For constructing the composed-query anchor $Q_{\text{comp}}$, we set the Slerp interpolation ratio $\lambda$ to 0.8 for CIRCO and Fashion-IQ, and 0.9 for CIRR, following the setting of a previous method~\cite{TAT}.
For TSF, the fusion weights are assigned to the fine-grained representation, coarse-grained representation, and composed-query anchor, respectively.
Specifically, in the order of $(T_{\text{fine}},T_{\text{coarse}},Q_{\text{comp}})$, we set the weights to $(0.3,0.4,0.3)$ for CIRCO, $(0.2,0.3,0.5)$ for CIRR, and $(0.3,0.1,0.6)$ for Fashion-IQ.
The number of Karcher mean iterations is fixed to 2 for all datasets.

\begin{table}[!th]
\centering
\caption{\textbf{Comparison with the state-of-the-art methods on Fashion-IQ.} R$_{\text{mean}}$ represents the average results across all the metrics. Best results are highlighted in bold, and the second best are underlined.}
\label{table:fashioniq}
\resizebox{0.95\linewidth}{!}{
\begin{tabular}{llccccccccccccc}
\Xhline{2\arrayrulewidth}
\noalign{\smallskip}

\multirow{2.7}{*}{\textbf{Backbone}} & \multirow{2.7}{*}{\textbf{Method}} & \multirow{2.7}{*}{\shortstack{\textbf{Training}\\ \textbf{-free}}} & \multicolumn{2}{c}{\textbf{Shirt}} & & \multicolumn{2}{c}{\textbf{Dress}} & & \multicolumn{2}{c}{\textbf{Tops\&Tee}} & & \multicolumn{3}{c}{\textbf{Avg.}} \\

\noalign{\smallskip}
\cline{4-5}
\cline{7-8}
\cline{10-11}
\cline{13-15}
\noalign{\smallskip}

 & &  & R@10 & R@50 & & R@10 & R@50 & & R@10 & R@50 & & R@10 & R@50 & R\textsubscript{mean} \\

\noalign{\smallskip}
\Xhline{2\arrayrulewidth}
\noalign{\smallskip}

\multirow{9}{*}{ViT-B/32}
 & PALAVRA \cite{PALAVRA} & \xmark & 21.49 & 37.05 & & 17.25 & 35.94 & & 20.55 & 38.76 & & 19.76 & 37.25 & 28.50 \\
 & SEARLE \cite{CIRCO} & \xmark & 24.44 & 41.61 & & 18.54 & 39.51 & & 25.70 & 46.46 & & 22.89 & 42.53 & 32.71 \\
 & Slerp \cite{TAT} & \cmark & 20.53 & 41.00 & & 23.75 & 40.92 & &  26.98 & 46.77 & & 23.75 & 42.90 & 33.33 \\
 & CIReVL \cite{CIReVL} & \cmark & 28.36 & 47.84 & & 25.29 & 46.36 & &  31.21 & 53.85 & & 28.29 & 49.35 & 38.82 \\
 & LDRE \cite{LDRE} & \cmark & 27.38 & 46.27 & & 19.97 & 41.84 & & 27.07 & 48.78 & & 24.81 & 45.63 & 35.22 \\
 & SEIZE \cite{SEIZE} & \cmark & 29.38 & 47.97 & & 25.37 & 46.84 & & 32.07 & 54.78 & & 28.94 & 49.86 & 39.40\\
 & OSrCIR~\cite{OSrCIR} & \cmark & 31.16 & 51.13 & & 29.35 & 50.37 & & 36.51 & 58.71 & & 32.34 & 53.40 & 42.87\\
 & CoTMR \cite{CoTMR} & \cmark & \underline{33.42} & \underline{53.93} & & \underline{31.09} & \underline{54.54} & & \underline{38.40} & \underline{61.14} & & \underline{34.30} & \underline{56.54} & \underline{45.42} \\
 & \baseline{\textbf{\modelname\ (Ours)}} & \baseline{\cmark} & \baseline{\textbf{38.35}} & \baseline{\textbf{58.40}} & \baseline{} & \baseline{\textbf{37.79}} & \baseline{\textbf{60.63}} & \baseline{} & \baseline{\textbf{45.80}} & \baseline{\textbf{66.99}} & \baseline{} & \baseline{\textbf{40.65}} & \baseline{\textbf{62.01}} & \baseline{\textbf{51.33}} \\

\noalign{\smallskip}
\hline
\noalign{\smallskip}

\multirow{8}{*}{ViT-L/14}
 & SEARLE \cite{CIRCO} & \xmark & 26.89 & 45.58 & & 20.48 & 43.13 & & 29.32 & 49.97 & & 25.56 & 46.23 & 35.89 \\
 & Slerp \cite{TAT} & \cmark & 21.96 & 41.55 & & 28.66 & 43.96 & &  30.24 & 48.34 & & 26.95 & 44.62 & 35.79 \\ 
 & CIReVL \cite{CIReVL} & \cmark & 29.49 & 47.40 & & 24.79 & 44.76 & & 31.36 & 53.65 & & 28.55 & 48.57 & 38.56 \\
 & LDRE \cite{LDRE} & \cmark & 31.04 & 51.22 & & 22.93 & 46.76 & & 31.57 & 53.64 & & 28.51 & 50.54 & 39.52 \\
 & SEIZE \cite{SEIZE} & \cmark & 33.04 & 53.22 & & 30.93 & 50.76 & & 35.57 & 58.64 & & 33.18 & 54.21 & 43.69\\
 & OSrCIR~\cite{OSrCIR} & \cmark & 33.17 & 52.03 & & 29.70 & 51.81 & & 36.92 & 59.27 & & 33.26 & 54.37 & 43.82\\ 
 
 & CoTMR \cite{CoTMR} & \cmark & \underline{35.43} & \underline{54.91} & & \underline{31.18} & \underline{55.04} & & \underline{38.55} & \underline{61.33} & & \underline{35.05} & \underline{57.09} & \underline{46.50} \\
 & \baseline{\textbf{\modelname\ (Ours)}} & \baseline{\cmark} & \baseline{\textbf{40.10}} & \baseline{\textbf{58.14}} & \baseline{} & \baseline{\textbf{37.87}} & \baseline{\textbf{59.65}} & \baseline{} & \baseline{\textbf{46.15}} & \baseline{\textbf{67.19}} & \baseline{} & \baseline{\textbf{41.37}} & \baseline{\textbf{61.66}} & \baseline{\textbf{51.52}} \\

\noalign{\smallskip}
\hline
\noalign{\smallskip}

\multirow{6}{*}{ViT-G/14}
 & CIReVL \cite{CIReVL} & \cmark & 33.71 & 51.42 & & 27.07 & 49.53 & & 35.80 & 56.14 & & 32.19 & 52.36 & 42.27 \\
 & LDRE \cite{LDRE} & \cmark & 35.94 & 58.58 & & 26.11 & 51.12 & & 35.42 & 56.67 & & 32.49 & 55.46 & 43.97 \\
 & SEIZE \cite{SEIZE} & \cmark & \underline{43.60} & \textbf{65.42} & & \underline{39.61} & \underline{61.02} & & \underline{45.94} & \textbf{71.12} & & \underline{43.05} & \textbf{65.85} & \underline{54.45}\\
 & OSrCIR~\cite{OSrCIR} & \cmark & 38.65 & 54.71 & & 33.02 & 54.78 & & 41.04 & 61.83 & & 37.57 & 57.11 & 47.34\\
 & CoTMR~\cite{CoTMR} & \cmark & 38.32 & 62.24 & & 34.51 & 57.36 & & 41.90 & 64.30 & & 38.25 & 61.32 & 49.78 \\
 & \baseline{\textbf{\modelname\ (Ours)}} & \baseline{\cmark} & \baseline{\textbf{45.17}} & \baseline{\underline{62.75}} & \baseline{} & \baseline{\textbf{42.00}} & \baseline{\textbf{63.56}} & \baseline{} & \baseline{\textbf{49.86}} & \baseline{\underline{70.91}} & \baseline{} & \baseline{\textbf{45.68}} & \baseline{\underline{65.74}} & \baseline{\textbf{55.71}} \\

\noalign{\smallskip}
\Xhline{2\arrayrulewidth}

\end{tabular}
}

\end{table}

\subsection{Main Results}
\label{subsec:main_results}

\noindent \textbf{CIRCO.}
Table~\ref{table:main} shows the performance comparison on CIRCO, which measures retrieval performance using mAP.
Unlike R@K, mAP reflects not only the accuracy of top-ranked results but also the overall ordering quality across all retrieved items, providing a more comprehensive evaluation of retrieval consistency.
Across all backbones, our method achieves the highest mAP, showing a substantial improvement over the existing training-free and fine-tuned approaches.
These consistent improvements indicate that combining dual-level target descriptions with the composed-query anchor helps capture target semantics more robustly and stabilizes the overall retrieval ranking.

\noindent \textbf{CIRR.}
Table~\ref{table:main} also shows the comparison with state-of-the-art methods on CIRR.
Our method achieves strong performance across most evaluation metrics, demonstrating its effectiveness on natural-image composed retrieval.
While our method obtains slightly lower scores on several ViT-G/14 settings, it remains competitive with existing training-free methods.
R\textsubscript{sub}@K focuses on a small curated subset based on cosine similarity using ResNet~\cite{ResNet}, which emphasizes appearance-based similarity.
The overall performance indicates that combining dual-level target descriptions with the composed-query anchor effectively captures target semantics beyond appearance-level similarity, demonstrating its robustness in real-world retrieval settings.

\noindent \textbf{Fashion-IQ.}
Table~\ref{table:fashioniq} reports the comparison with state-of-the-art methods on Fashion-IQ with CLIP.
Our method achieves the best overall performance in terms of R\textsubscript{mean}, demonstrating its effectiveness on fashion-domain composed retrieval.
Although our method shows slightly lower performance than SEIZE~\cite{SEIZE} on a few R@50 metrics with the ViT-G/14 backbone, it still achieves the highest overall R\textsubscript{mean}.
Moreover, our method consistently improves as the backbone capacity increases from ViT-B/32 to ViT-G/14, indicating its scalability and robustness across different CLIP backbones.

\subsection{Further Analysis}
\label{subsec:further_analysis}

\begin{table}[!tb]
\centering
\caption{\textbf{Ablation study with CLIP ViT-G/14.}
FD, CD, and Comp denote fine-grained descriptions, coarse-grained descriptions, and the composed-query anchor, respectively.
The last row corresponds to the full \modelname.}
\label{table:ablation}
\setlength{\tabcolsep}{4.0pt}
\resizebox{0.8\linewidth}{!}{
\begin{tabular}{ccc cc cc cc}
\Xhline{2\arrayrulewidth}
\noalign{\smallskip}

\multirow{2}{*}{\textbf{FD}} &
\multirow{2}{*}{\textbf{CD}} &
\multirow{2}{*}{\textbf{Comp}} &
\multicolumn{2}{c}{\textbf{CIRCO}} &
\multicolumn{2}{c}{\textbf{CIRR}} &
\multicolumn{2}{c}{\textbf{Fashion-IQ}} \\

\cmidrule(lr){4-5}
\cmidrule(lr){6-7}
\cmidrule(lr){8-9}

& & &
\textbf{mAP@5} & \textbf{mAP@10} &
\textbf{R@5} & \textbf{R\textsubscript{sub}@1} &
\textbf{R@10} & \textbf{R@50} \\

\noalign{\smallskip}
\Xhline{2\arrayrulewidth}
\noalign{\smallskip}

\cmark &        &        & 25.55 & 26.66 & 60.00 & 61.54 & 35.57 & 56.03 \\
       & \cmark &        & 32.03 & 33.34 & 64.29 & 69.06 & 32.83 & 52.84 \\
       &        & \cmark & 17.32 & 17.88 & 62.75 & 67.42 & 40.58 & 60.39 \\

\noalign{\smallskip}
\hline
\noalign{\smallskip}

\cmark & \cmark &        & 34.77 & 35.95 & 67.04 & 68.15 & 38.43 & 59.35 \\
\cmark &        & \cmark & 32.01 & 32.88 & 69.33 & 67.66 & 43.86 & 65.23 \\
       & \cmark & \cmark & 34.31 & 34.99 & 69.74 & \textbf{70.58} & 41.49 & 63.22 \\

\noalign{\smallskip}
\hline
\noalign{\smallskip}

\cmark & \cmark & \cmark & \textbf{37.47} & \textbf{38.63} & \textbf{71.40} & 70.27 & \textbf{45.68} & \textbf{65.74} \\

\noalign{\smallskip}
\Xhline{2\arrayrulewidth}
\end{tabular}
}
\end{table}

\noindent \textbf{Ablation study on DCR-TSF.}
Table~\ref{table:ablation} shows the ablation results of the three components in DCR-TSF across CIRCO, CIRR, and Fashion-IQ using CLIP ViT-G/14.
In this experiment, FD, CD, and Comp indicate the fine-grained representation $T_{\text{fine}}$, the coarse-grained representation $T_{\text{coarse}}$, and the composed-query anchor $Q_{\text{comp}}$, respectively.
The single-component results reveal that the most effective retrieval cue differs depending on the dataset characteristics.
On open-domain benchmarks such as CIRCO and CIRR, CD generally provides strong performance, suggesting that coarse semantic abstraction is useful when queries involve diverse scenes and multiple plausible target images.
In contrast, on Fashion-IQ, FD and Comp are more effective than CD, indicating that fashion-domain retrieval relies more heavily on fine-grained visual attributes and direct preservation of the reference-modification query.
The pairwise combinations further demonstrate the complementarity among the three components.
Combining FD and CD balances detailed visual evidence with coarse semantic abstraction, while combining Comp with FD or CD provides information that is not fully captured by MLLM-generated descriptions.
Finally, using all three components achieves the best performance on most metrics, confirming that DCR-TSF benefits from jointly exploiting fine-grained visual evidence, coarse-grained semantic abstraction, and direct preservation of the original composed query.

\begin{figure}[!t]
    \centering

    \begin{subfigure}[t]{0.32\linewidth}
        \centering
        \includegraphics[width=\linewidth]{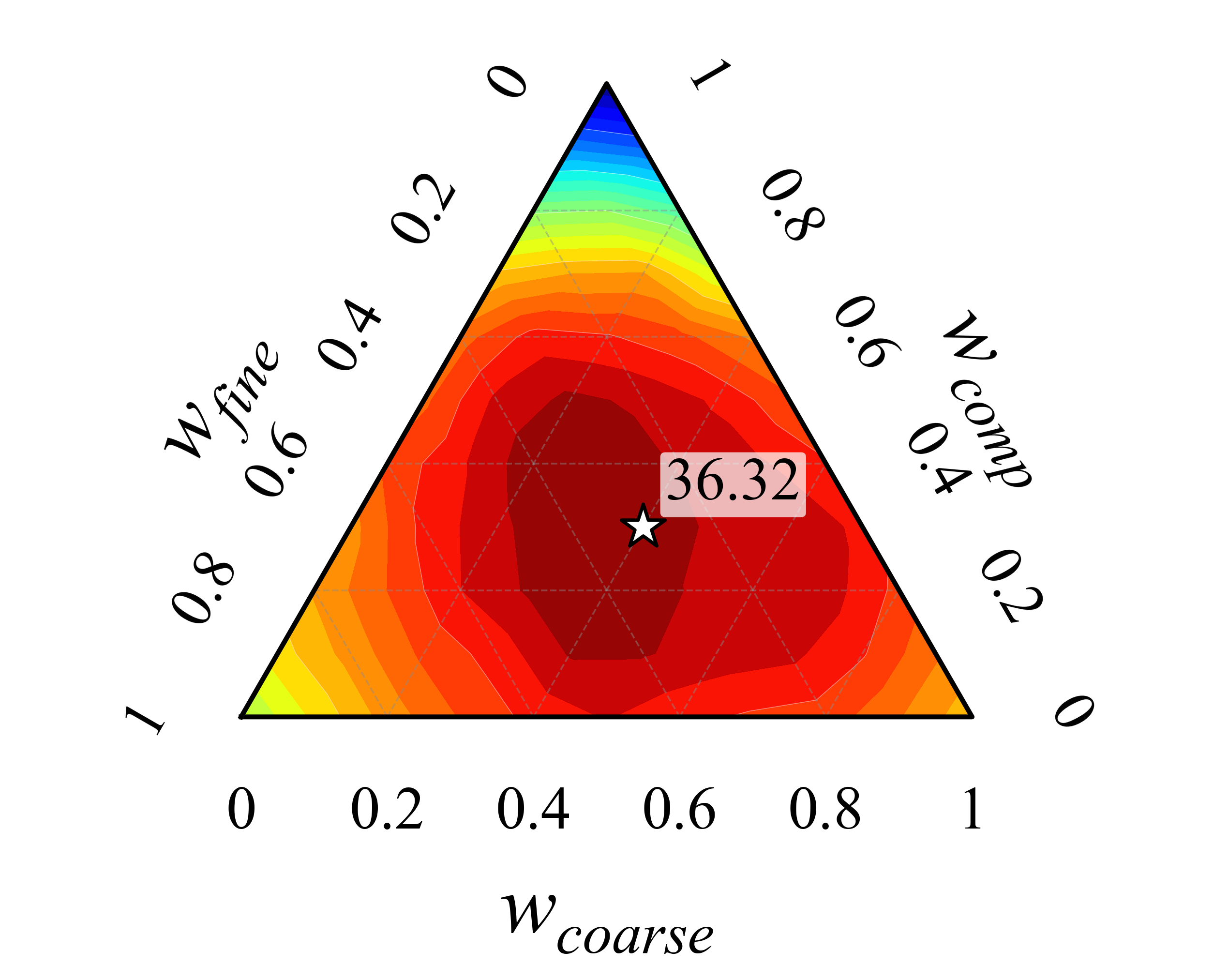}
        \vspace{-12pt}
        \caption{CIRCO}
        \label{subfig:beta_gamma_circo}
    \end{subfigure}%
    \hspace{4pt}
    \begin{subfigure}[t]{0.32\linewidth}
        \centering
        \includegraphics[width=\linewidth]{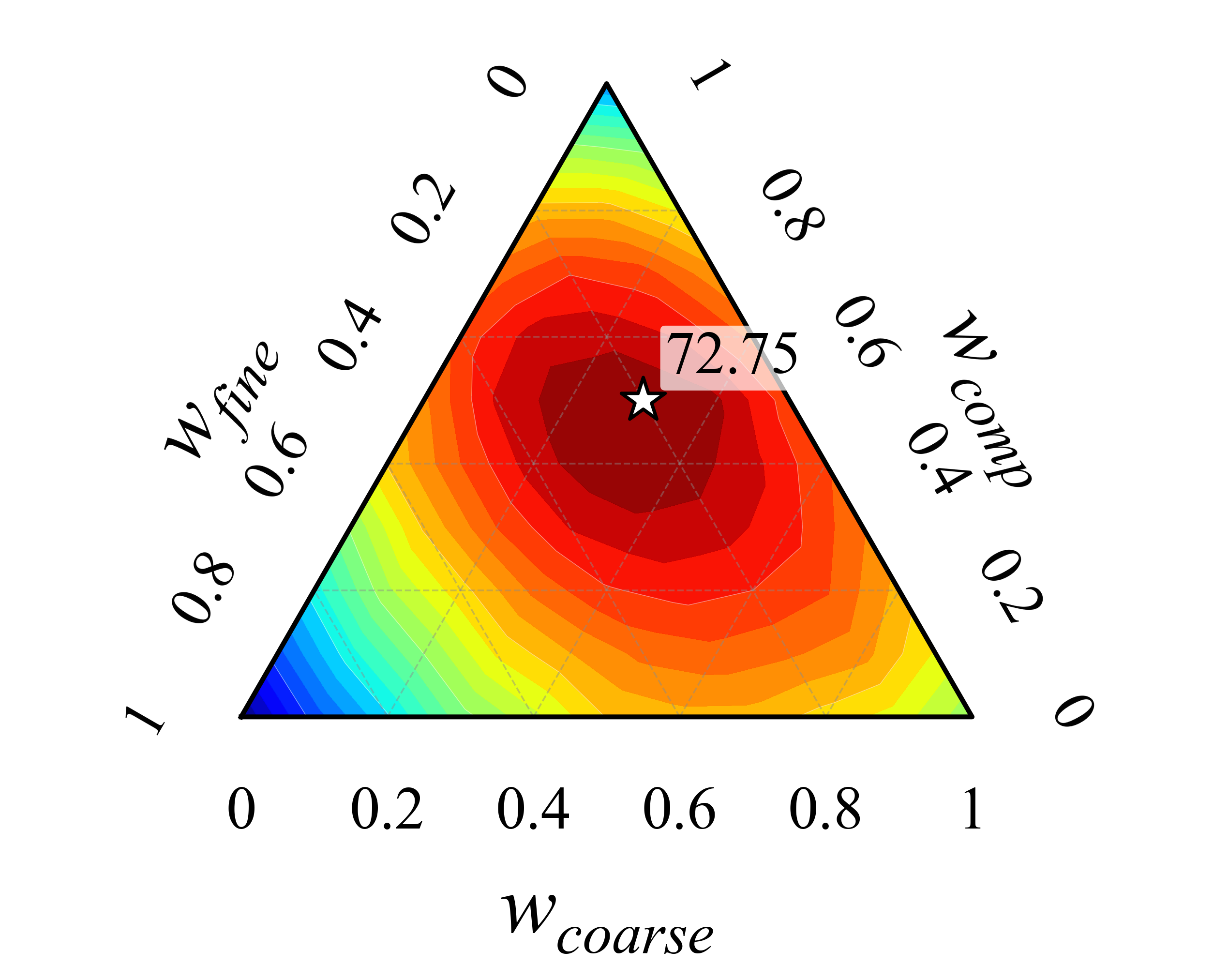}
        \vspace{-12pt}
        \caption{CIRR}
        \label{subfig:beta_gamma_cirr}
    \end{subfigure}%
    \hspace{4pt}
    \begin{subfigure}[t]{0.32\linewidth}
        \centering
        \includegraphics[width=\linewidth]{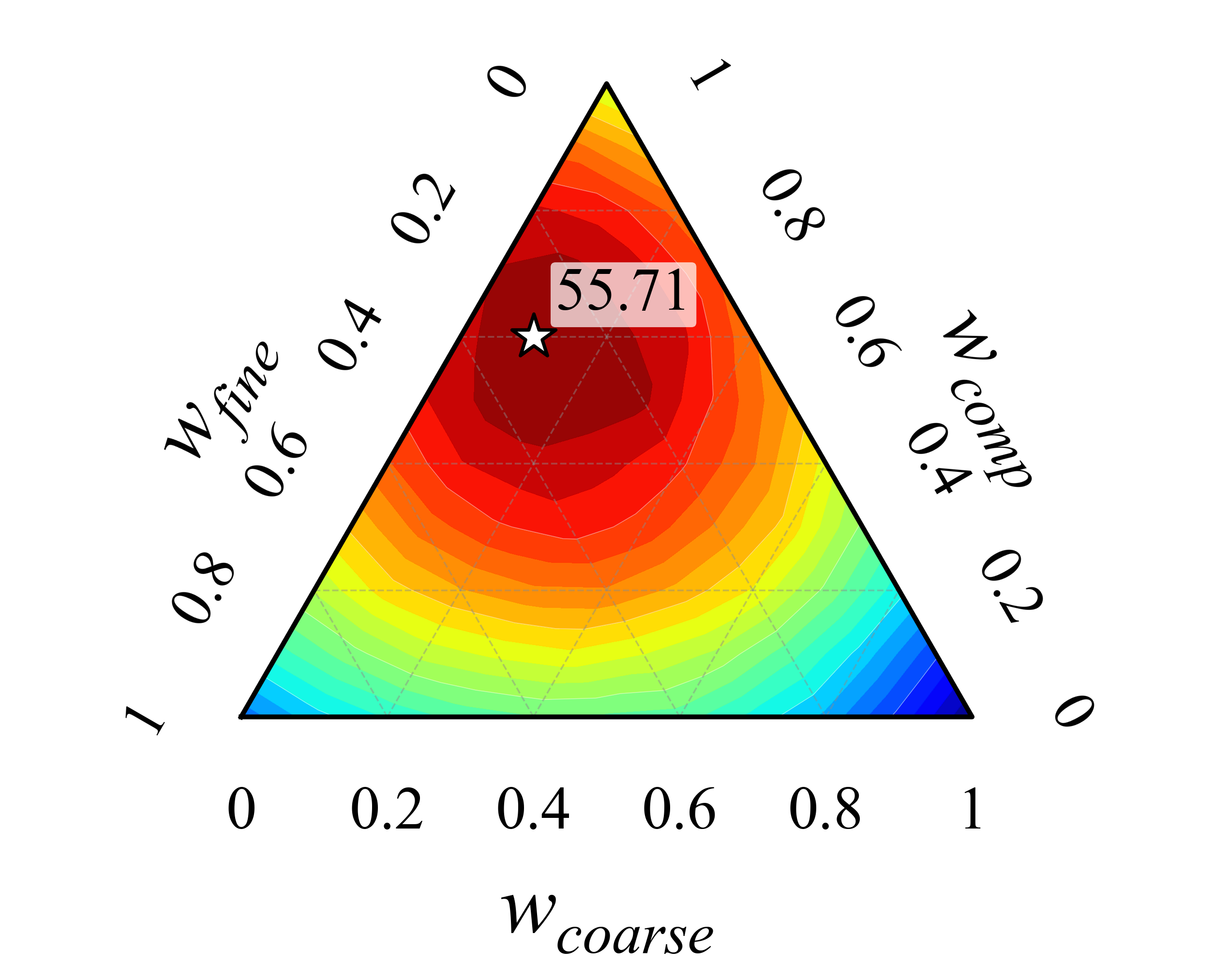}
        \vspace{-12pt}
        \caption{Fashion-IQ}
        \label{subfig:beta_gamma_fiq}
    \end{subfigure}%

    \caption{\textbf{Hyperparameter sensitivity of TSF fusion weights.}
    Ternary plots show validation results for different TSF fusion-weight configurations on each dataset using CLIP ViT-G/14.
    The star marker indicates the best-performing configuration selected for the main experiments.}
    \label{fig:beta_gamma}
    \vspace{-12pt}
\end{figure}

\subsection{Hyperparameter Sensitivity}
\label{subsec:hyperparameter_sensitivity}
Fig.~\ref{fig:beta_gamma} shows the sensitivity analysis of TSF fusion weights using ternary plots on the validation splits of CIRCO, CIRR, and Fashion-IQ.
We vary the fusion weights $\mathbf{w}=(w_{\text{fine}},w_{\text{coarse}},w_{\text{comp}})$ under the constraint $w_{\text{fine}}+w_{\text{coarse}}+w_{\text{comp}}=1$, and report the retrieval performance for each weight combination.
Each point in the ternary plot corresponds to a specific weight configuration, and the star marker indicates the best-performing configuration used in our main experiments:
$\mathbf{w}=(0.3,0.4,0.3)$ for CIRCO, $\mathbf{w}=(0.2,0.3,0.5)$ for CIRR, and $\mathbf{w}=(0.3,0.1,0.6)$ for Fashion-IQ.
Overall, TSF is not highly sensitive to a precise weight configuration.
Although the best-performing weights differ across datasets, high-performance regions are broadly formed around balanced combinations of the three components, especially near the center of the ternary plots.
This suggests that TSF provides stable performance as long as fine-grained descriptions, coarse-grained descriptions, and the composed-query anchor are jointly considered.
On CIRCO, strong performance is obtained when the three components are well balanced, indicating that open-domain retrieval benefits from both semantic abstraction and original query preservation.
On CIRR and Fashion-IQ, relatively larger weights for the composed-query anchor are preferred, reflecting the importance of preserving the original reference-modification information in natural-image retrieval with visually similar candidates and fashion-domain retrieval.
These results support the role of TSF in balancing fine-grained visual evidence, coarse-grained semantic abstraction, and direct preservation of the original composed query.

\begin{figure*}[t]
    \centering

    \begin{minipage}[c]{0.48\textwidth}
        \centering
        \captionof{table}{\textbf{Disjointness analysis on CIRCO with ViT-G/14.}
        Jaccard and Overlap Coefficient are computed between Top-$K$ retrieval sets from fine- and coarse-level representations.}
        \label{table:jaccard}
        
        \resizebox{0.85\linewidth}{!}{
        \begin{tabular}{c cc}
        \Xhline{2\arrayrulewidth}
        \noalign{\smallskip}

        Top-$K$ & Jaccard & Overlap Coefficient \\

        \noalign{\smallskip}
        \Xhline{2\arrayrulewidth}
        \noalign{\smallskip}

        5  & 0.3508 & 0.4237 \\
        10 & 0.4140 & 0.5050 \\
        25 & 0.5350 & 0.6550 \\
        50 & 0.6355 & 0.7733 \\

        \noalign{\smallskip}
        \Xhline{2\arrayrulewidth}
        \end{tabular}
        }
    \end{minipage}%
    \hfill
    \begin{minipage}[c]{0.48\textwidth}
        \centering
        \vspace{8pt}
        \includegraphics[width=0.78\linewidth]{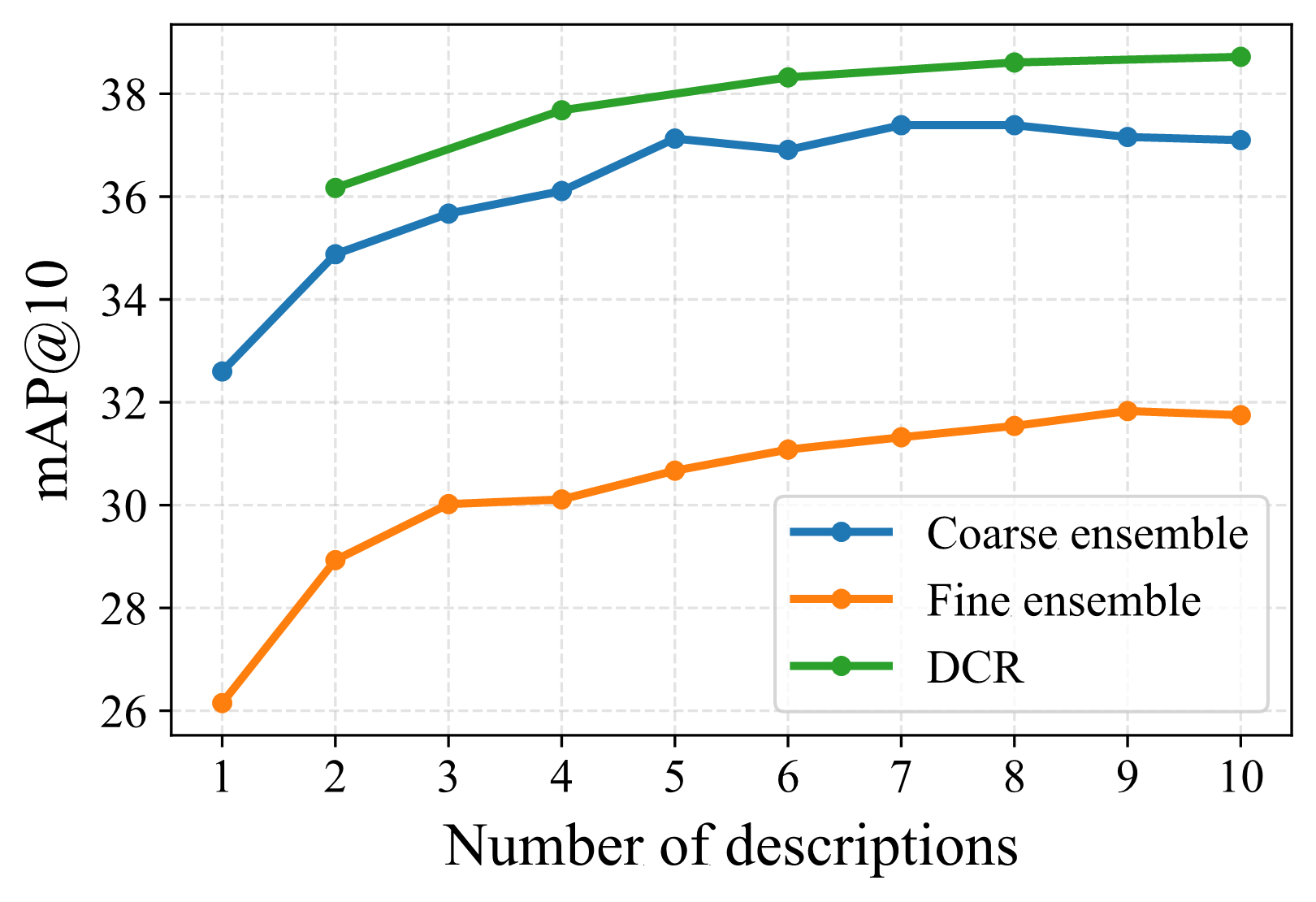}

        \captionof{figure}{\textbf{Comparison of caption ensemble and DCR on the CIRCO test set with ViT-G/14.}}
        \label{fig:caption_ensemble}
    \end{minipage}
    \vspace{-15pt}
\end{figure*}

\noindent \textbf{Disjointness Analysis of Fine- and Coarse-Descriptions.}
Table~\ref{table:jaccard} shows the Jaccard similarity and overlap coefficient, which measure the degree of intersection between the Top-$K$ retrieval sets obtained from the fine- and coarse-grained descriptions on CIRCO.
For each reference image, we retrieve the Top-$K$ images using both the fine- and coarse- descriptions, compare the overlap between the two corresponding retrieval images within ground truth, and then average the results over all queries.
The Top-K overlap between fine- and coarse-grained retrieval sets is low at small $K$, implying that the two description levels capture distinct information.  
As $K$ increases, the overlap ratio gradually rises, suggesting that fine- and coarse-descriptions share a consistent high-level concept.
This trend indicates that the two reasoning levels follow similar semantic information but emphasize different aspects of the reference image and modification text.

\noindent \textbf{Comparative Analysis of Caption Ensemble and DCR.}
Existing methods~\cite{LDRE,IP-CIR,SEIZE} report that utilizing multiple generated captions can improve the performance, raising the possibility that the gains of our DCR may stem from caption ensemble. 
To investigate this hypothesis, we generate multiple descriptions with varying temperature using either $P_{\text{fine}}$ or $P_{\text{coarse}}$ independently and compare their performance with DCR under the same number of descriptions.
Fig.~\ref{fig:caption_ensemble} shows that increasing the number of fine- or coarse-only descriptions yields only marginal improvements and quickly saturates.
In contrast, DCR consistently achieves superior performance when using the same number of descriptions, indicating that the performance gain does not arise from simple caption ensemble.
These results suggest that DCR benefits from complementary semantic granularity, where fine- and coarse-grained descriptions provide mutually informative cues for compositional understanding.
Although performance generally increases as more descriptions are used, generating multiple captions requires repeated MLLM inference, introducing substantial computational overhead.
Therefore, in all experiments, we employ one fine- and one coarse-grained description for DCR.

\noindent \textbf{Analysis on Scalability and Model Variation in MLLMs.}
Table~\ref{table:mllm} shows the retrieval performance of various MLLMs with different parameter scales on CIRCO.
As shown in Table~\ref{table:mllm}, larger MLLMs generally achieve higher retrieval performance, but the improvement becomes marginal beyond 32B parameters.
Specifically, Qwen2.5-VL-32B achieves a balance between performance and efficiency, while further scaling to 72B provides only marginal gains.
We also report results with the latest Qwen3-VL series and GPT-5, which show comparable performance to Qwen2.5-VL-32B.
This suggests that our framework is not highly sensitive to a specific MLLM backbone among sufficiently capable models.
However, to ensure fair comparison and maintain consistency with previous studies, we adopt Qwen2.5-VL-32B for all experiments.















\begin{table}[!t]
\centering
\caption{\textbf{Analysis on scalability and model variation in MLLMs on the CIRCO test set with CLIP ViT-G/14.}}
\label{table:mllm}
\vspace{-5pt}
\resizebox{0.64\linewidth}{!}{
\begin{tabular}{
l
>{\hspace{3pt}}c<{\hspace{3pt}}
>{\hspace{3pt}}c<{\hspace{3pt}}
>{\hspace{3pt}}c<{\hspace{3pt}}
>{\hspace{3pt}}c<{\hspace{3pt}}
}

\Xhline{2\arrayrulewidth}
\noalign{\smallskip}

\textbf{Method} & mAP@5 & mAP@10 & mAP@25 & mAP@50 \\

\noalign{\smallskip}
\Xhline{2\arrayrulewidth}
\noalign{\smallskip}

Qwen2.5-VL-3B~\cite{Qwen2.5} & 23.72 & 24.89 & 27.09 & 28.09\\
Qwen2.5-VL-7B~\cite{Qwen2.5} & 30.62 & 31.86 & 34.29 & 35.42\\
\baseline{Qwen2.5-VL-32B~\cite{Qwen2.5}} & \baseline{34.77} & \baseline{35.95} & \baseline{38.56} & \baseline{39.79}\\
Qwen2.5-VL-72B~\cite{Qwen2.5} & \textbf{35.26} & \textbf{36.69} & \textbf{39.46} & \textbf{40.62}\\

\noalign{\smallskip}
\hline
\noalign{\smallskip}

Qwen2-VL-72B~\cite{Qwen2} & 34.38 & 36.11 & 38.81 & 39.89\\

\noalign{\vspace{1.2pt}}
\hdashline
\noalign{\vspace{1.2pt}}

Qwen3-VL-4B~\cite{Qwen3} & 33.21 & 34.48 & 37.23 & 38.37\\
Qwen3-VL-8B~\cite{Qwen3} & {34.81} & {35.87} & {38.50} & {39.60}\\
Qwen3-VL-32B~\cite{Qwen3} & 34.49 & 35.72 & {38.35} & {39.48}\\

\noalign{\vspace{1.2pt}}
\hdashline
\noalign{\vspace{1.2pt}}

GPT-5~\cite{GPT-5} & 35.18 & 36.68 & 39.49 & 40.64\\

\noalign{\smallskip}
\Xhline{2\arrayrulewidth}

\end{tabular}
}
\vspace{-5pt}
\end{table}

\begin{table}[!t]
\centering
\caption{\textbf{Inference time comparison with existing training-free ZS-CIR methods.}
We report the average latency per composed query in seconds.
Each column decomposes the total inference time into the main reasoning stages of each method.}
\vspace{-5pt}
\label{table:inference_time}
\resizebox{0.85\linewidth}{!}{
\begin{tabular}{l@{\hspace{6pt}} c@{\hspace{6pt}}c@{\hspace{6pt}}c}
\Xhline{2\arrayrulewidth}
\noalign{\smallskip}

BLIP-2 \& GPT-3.5-turbo & Multi-caption generator & LLM-based editing reasoner & Total\\
\noalign{\smallskip}
\hline
\noalign{\smallskip}
SEIZE~\cite{SEIZE} & 1.560 $\pm$ 0.056  & 3.912 $\pm$ 0.812 & 5.472 $\pm$ 0.791\\
\noalign{\smallskip}
\Xhline{2\arrayrulewidth}
\noalign{\smallskip}
Qwen2.5-VL-32B & Image-scale reasoning & Object-scale reasoning & Total\\
\noalign{\smallskip}
\hline
\noalign{\smallskip}
CoTMR~\cite{CoTMR} & 1.892 $\pm$ 0.011 & 1.661 $\pm$ 0.014 & 3.553 $\pm$ 0.023\\

\noalign{\smallskip}
\Xhline{2\arrayrulewidth}
\noalign{\smallskip}
Qwen2.5-VL-32B & Fine-grained description & Coarse-grained description & Total\\
\noalign{\smallskip}
\hline
\noalign{\smallskip}
DCR-TSF (Ours) & 1.883 $\pm$ 0.019 & 1.584 $\pm$ 0.018 & 3.467 $\pm$ 0.030\\

\noalign{\smallskip}
\Xhline{2\arrayrulewidth}
\end{tabular}
}
\vspace{-10pt}
\end{table}

\noindent \textbf{Computational Cost Analysis.}
We analyze the computational overhead of our method when processing a single composed query.
The overall latency is mainly dominated by MLLM inference for generating fine- and coarse-grained descriptions, while the remaining retrieval operations introduce only marginal additional cost.
In particular, TSF only performs Slerp-based anchor construction and weighted Karcher mean fusion in the CLIP embedding space, introducing negligible computational overhead compared to MLLM inference.
Table~\ref{table:inference_time} compares the inference time of our method with SEIZE~\cite{SEIZE} and CoTMR~\cite{CoTMR} under the same experimental setting.
Under the same environment with an NVIDIA A6000 GPU, our method requires 3.467$\pm$0.030 seconds per query.
This is slightly faster than CoTMR and substantially more efficient than SEIZE, which require 3.553$\pm$0.023 and 5.472$\pm$0.791 seconds per query, respectively.
Specifically, generating the fine- and coarse-grained descriptions takes 1.883$\pm$0.019 and 1.584$\pm$0.018 seconds, respectively, showing that the computational cost mainly comes from MLLM-based description generation.
Despite this overhead, the dual-level reasoning stage provides complementary semantic cues, and TSF effectively integrates them with the original composed-query information.
These results indicate that our method achieves a favorable balance between retrieval performance and computational efficiency.

\begin{figure}[!t]
    \centering
    
    \begin{subfigure}[t]{0.83\linewidth}
        \centering
        \vspace{-2pt}
        \includegraphics[width=\linewidth]{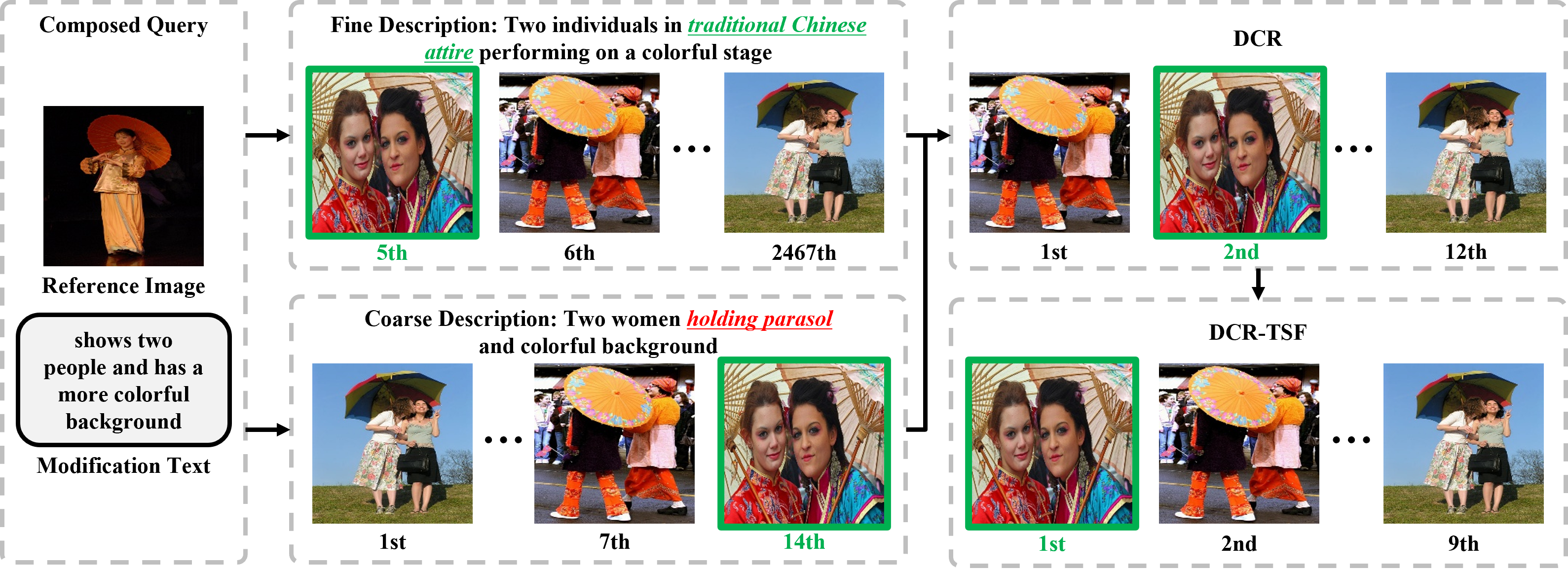}
        \vspace{-10pt}
    \end{subfigure}%
    \\
    \begin{subfigure}[t]{0.83\linewidth}
        \centering
        \includegraphics[width=\linewidth]{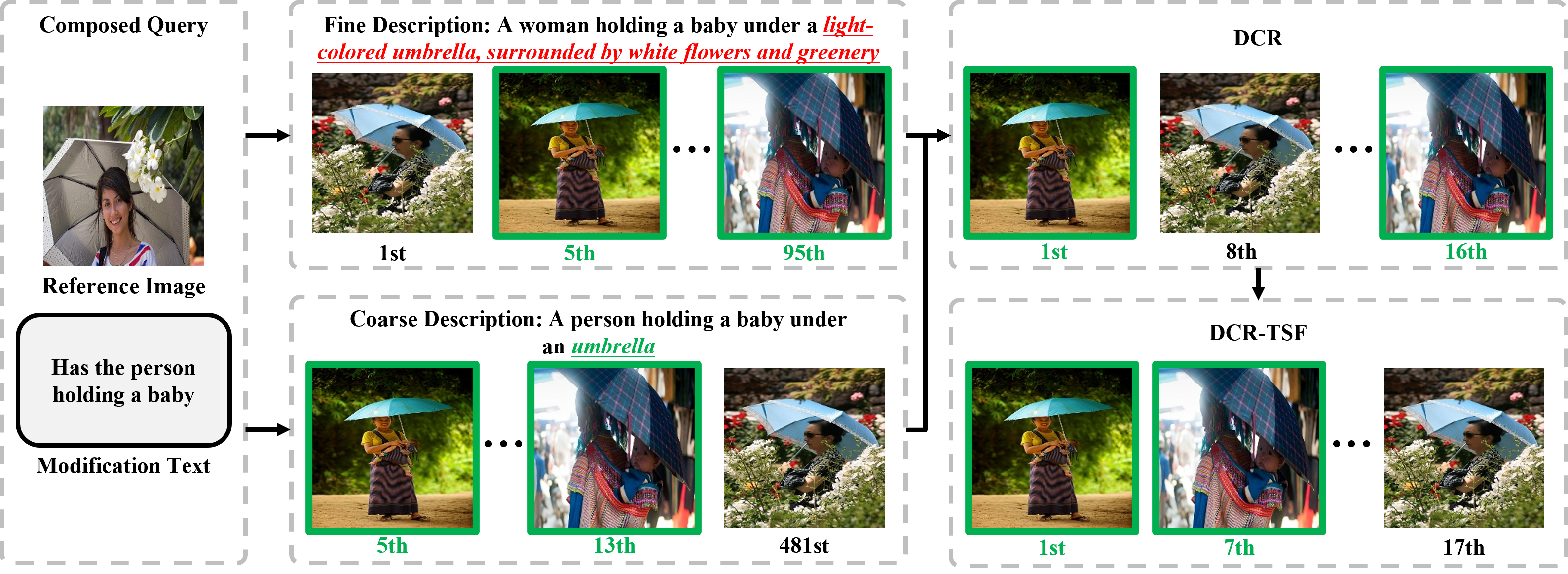}
        \vspace{-7pt}
    \end{subfigure}%
    \vspace{-7pt}
    \caption{\textbf{Qualitative comparison on CIRCO with ViT-G/14.}
    Each example shows a reference image, its modification text, and the retrieved images with their ranks under four settings: \emph{Fine-Grained Description}, \emph{Coarse-Grained Description}, \emph{DCR}, and \emph{\modelname}.
    The ground-truth image is highlighted in green.}
    \label{fig:ret_results}
\vspace{-10pt}
\end{figure}
\subsection{Qualitative Results}
\label{subsec:qualitative_results}
Fig.~\ref{fig:ret_results} presents qualitative retrieval examples on CIRCO.
The ground-truth images highlighted in green often appear at suboptimal ranks when relying solely on either fine- or coarse-grained descriptions.
DCR improves the ranking by combining fine-grained visual details with coarse-grained semantic cues.
By further incorporating the composed-query anchor, \modelname\ produces more balanced retrieval results and raises the ranks of the ground-truth images.
\section{Conclusion}
\label{sec:conclusion}
In this paper, we propose \modelname, a training-free framework for ZS-CIR that combines Dual-level Compositional Reasoning (DCR) and Triadic Spherical Fusion (TSF).
Our method generates fine- and coarse-grained descriptions to capture complementary semantics, balancing detailed visual evidence with high-level semantic abstraction.
To preserve the original image-text query information, TSF constructs a composed-query anchor from the reference image and modification text and fuses it with the fine- and coarse-grained representations on the unit hypersphere.
By exploiting fine-grained visual evidence, coarse-grained semantic abstraction, and direct preservation of the original composed query, \modelname\ mitigates the limitations of relying on fine-grained generated descriptions alone.
Experiments on CIRCO, CIRR, and Fashion-IQ demonstrate that \modelname\ achieves state-of-the-art performance among training-free ZS-CIR methods.


\subsubsection*{Acknowledgments.}
This work was supported by the Institute of Information \& Communications Technology Planning \& Evaluation (IITP) grant funded by the Korea government (MSIT) (No. RS-2019-II190079, Artificial Intelligence Graduate School Program, Korea University), and by the Information Technology Research Center (ITRC) support program (No. IITP-2026-RS-2024-00436857).
\newpage

\bibliographystyle{splncs04}
\bibliography{main}

\newpage
\clearpage

\renewcommand{\thesection}{\Alph{section}}
\renewcommand{\thefigure}{\Alph{figure}}
\renewcommand{\theequation}{\Alph{equation}}
\renewcommand{\thetable}{\Alph{table}}
\setcounter{section}{0}
\setcounter{table}{0}
\setcounter{figure}{0}
\setcounter{equation}{0}

\setcounter{page}{1}
\title{DCR-TSF: Dual-Level Compositional Reasoning with Triadic Spherical Fusion for Training-Free Zero-Shot Composed Image Retrieval\\ Supplementary}

\author{
Gwang-Ho Na\inst{1}\orcidlink{0009-0005-8543-8347} \and
Ho-Joong Kim\inst{1}\orcidlink{0000-0003-4200-5136} \and
Seong-Whan Lee\inst{1}\orcidlink{0000-0002-6249-4996}
}

\authorrunning{G.-H. Na et al.}

\titlerunning{DCR-TSF}

\institute{
Dept. of Artificial Intelligence, Korea University, Seoul, Korea\\
\email{\{g\_h\_na, hojoong\_kim, sw.lee\}@korea.ac.kr}
}

\maketitle

\section{Additional Details}
\label{sec:additional_details}

\subsection{Prompt for Fine-Coarse Description}
\noindent \textbf{Fine Description Prompt} $\mathbf{P_{\textbf{fine}}}$
Based on the existing methods~\cite{CoTMR,OSrCIR}, we design a fine-grained prompt that explicitly enumerates object attributes, spatial relations, and all visually grounded semantics in the reference image.
This prompt guides the model to focus on visual details of the scene.
Fig.~\ref{fig:prompt_fine} illustrates the full template.

\noindent \textbf{Coarse Description Prompt} $\mathbf{P_{\textbf{coarse}}}$
While $\mathbf{P_{\textbf{fine}}}$ focuses on detailed attributes, the coarse prompt provides a short summary that captures only the shared, global semantics of the reference image.
Fig.~\ref{fig:prompt_coarse} describes the coarse template.

\subsection{Preliminary on Triadic Spherical Fusion}
\label{subsec:tsf_details}

In this section, we provide additional details on the implementation of Triadic Spherical Fusion (TSF).
TSF fuses three unit-normalized representations: the fine-grained representation $T_{\text{fine}}$, the coarse-grained representation $T_{\text{coarse}}$, and the composed-query anchor $Q_{\text{comp}}$.
We denote this set of representations as
\begin{equation}
\mathcal{Z} = \{T_{\text{fine}}, T_{\text{coarse}}, Q_{\text{comp}}\}.
\end{equation}
Each representation lies on the unit hypersphere $\mathbb{S}^{d-1}$, where $d$ denotes the embedding dimension.
The fusion weight assigned to each $Z\in\mathcal{Z}$ is denoted as $w_Z$, with $\sum_{Z\in\mathcal{Z}}w_Z=1$.

\noindent \textbf{Weighted Karcher Mean.}
The Karcher mean, also known as the Riemannian center of mass, generalizes the Euclidean mean to Riemannian manifolds by minimizing the sum of squared geodesic distances~\cite{Karchermean}.
On the unit hypersphere, we define the weighted Karcher mean of $\mathcal{Z}$ as:
\begin{equation}
T_{\text{final}}
=
\arg\min_{Y \in \mathbb{S}^{d-1}}
\sum_{Z\in\mathcal{Z}}
w_Z d_{\mathbb{S}}^2(Y,Z),
\end{equation}
where $Y$ denotes an optimization variable on the unit hypersphere and $d_{\mathbb{S}}(\cdot,\cdot)$ denotes the geodesic distance on the unit hypersphere.
For two unit vectors $a$ and $b$, this distance is given by:
\begin{equation}
d_{\mathbb{S}}(a,b) = \cos^{-1}(a^\top b).
\end{equation}
This formulation differs from Euclidean averaging because it computes the mean with respect to the intrinsic geometry of the unit hypersphere.
Riemannian geometry and tangent-space mappings have also been employed for transfer learning and classification in other domains~\cite{1_prabhakar2023holistic}.

\noindent \textbf{Numerical Solver.}
The weighted Karcher mean generally does not have a simple closed-form solution for more than two points on the sphere.
Therefore, we compute it using a standard iterative procedure based on the logarithmic and exponential maps.
Given the current estimate $Y^{(r)}$, each representation $Z\in\mathcal{Z}$ is first mapped to the tangent space at $Y^{(r)}$:
\begin{equation}
\xi_Z^{(r)} = \log_{Y^{(r)}}(Z).
\end{equation}
The tangent vectors are then averaged using the fusion weights:
\begin{equation}
\Delta^{(r)}
=
\sum_{Z\in\mathcal{Z}}
w_Z \xi_Z^{(r)}.
\end{equation}
The estimate is updated by mapping the averaged tangent vector back to the unit hypersphere:
\begin{equation}
Y^{(r+1)}
=
\exp_{Y^{(r)}}(\Delta^{(r)}).
\end{equation}
After a fixed number of iterations, the final estimate is used as $T_{\text{final}}$.

\noindent \textbf{Sphere-specific Maps.}
For completeness, we describe the logarithmic and exponential maps used on the unit hypersphere.
Given two unit vectors $x$ and $y$, the logarithmic map at $x$ is:
\begin{equation}
\log_x(y)
=
\frac{\theta}{\sin \theta}
\left(
y - \cos\theta \, x
\right),
\quad
\theta = \cos^{-1}(x^\top y).
\end{equation}
Given a tangent vector $\xi$ at $x$, the exponential map is:
\begin{equation}
\exp_x(\xi)
=
\cos(\|\xi\|)x
+
\sin(\|\xi\|)
\frac{\xi}{\|\xi\|}.
\end{equation}
These maps allow TSF to perform weighted averaging intrinsically on the unit hypersphere rather than in the ambient Euclidean embedding space.

\noindent \textbf{Initialization and Stability.}
We initialize the iterative solver with the normalized weighted Euclidean average:
\begin{equation}
Y^{(0)}
=
\frac{
\sum_{Z\in\mathcal{Z}} w_Z Z
}{
\left\|
\sum_{Z\in\mathcal{Z}} w_Z Z
\right\|
}.
\end{equation}
For numerical stability, inner products used in $\cos^{-1}(\cdot)$ are clipped to a valid range.
When $\theta$ or $\|\xi\|$ is close to zero, we use the corresponding first-order approximation.
In practice, a small fixed number of iterations is sufficient for stable fusion.

\begin{figure}
\vspace{-5pt}
\centering
\begin{tcolorbox}[title={\textbf{$D_{fine}$}}]

\tiny

Your task is to modify the reference image based on the modification instructions and generate the updated image description. 
The description should be complete and can cover various semantic aspects, such as cardinality, addition, negation, direct addressing, compare \& change, comparative, conjunction, spatial relations \& background, viewpoint.

\medskip

\#\# Workflow
To complete the task accurately, please follow four steps:

\medskip

\#\#\# Step 1: Understand the Original Image

1. Identify all the objects, attributes, and their relationships in the image. 

2. Pay attention to the spatial relations, background, viewpoint in the image. 

3. Please complete this task step by step.

\medskip

\#\#\# Step 2: Analyze the Manipulation Text

1. Break down manipulation text into clear, actionable steps.

2. Classify each modification into semantic aspects:

\hspace*{1.5em}- Cardinality (e.g., changing number of objects)

\hspace*{1.5em}- Addition (e.g., adding new objects or elements)

\hspace*{1.5em}- Negation (e.g., removing objects or attributes)

\hspace*{1.5em}- Compare \& Change (e.g., swapping objects or styles)

\hspace*{1.5em}- Viewpoint (e.g., changing camera angle or scene perspective)

\hspace*{1.5em}- Conjunction (e.g., multiple simultaneous changes)

3. For each modification, identify what stays the same vs. what changes.

4. Please complete this task step by step.

\medskip

\#\#\# Step 3: Apply the Modifications

1. Apply all changes step-by-step while maintaining logical consistency.

2. Preserve context and relationships wherever possible.

3. Ensure the final scene remains visually coherent after modifications.

4. Based on the applied changes, write a preliminary descriptive sentence. This sentence will be polished in the final step.

5. Please complete this task step by step.

\medskip

\#\#\# Step 4: Generate the Target Image Description
Based on the sentence generated in Step 3, create a final, visual, and concise description that allows a user to immediately picture the image. 
This step is about refining a mechanical description into a natural, human-written photo caption.

1.  Reframe the sentence to focus on the image's main subject and its most important action or state.

2.  Omit subjective or unnecessarily detailed modifiers.

3.  Resolve all relative terms into concrete, explicit attribute values (e.g., ``the same color'' → ``blue'', ``more'' → ``five tables'').

4.  Condense the background into a single, representative noun that indicates the location, without listing detailed elements (e.g., ``a living room with a sofa, TV, and table'' → ``a living room''; ``a forest path with green trees'' → ``a forest path'').

5.  Avoid rigid descriptions; use a vivid and natural tone as if describing a photo you are looking at directly.

6.  The final description must be strictly one sentence only.

7.  Please complete this task step by step.

\#\# Input Format
\textless Input\textgreater

\{

\hspace*{1.5em}``Reference Image'': \textless image\_text\_or\_url\textgreater,

\hspace*{1.5em}``Modification Text'': \textless modification\_text\textgreater

\}

\textless /Input\textgreater

\medskip

\#\# Strict Output Format

\textless Response\textgreater

\{

\hspace*{1.5em}``Original Image Description'': ``\textless description\_of\_original\_image\textgreater'',

\hspace*{1.5em}``Thoughts'': ``\textless step-by-step reasoning on manipulation intents\textgreater'',

\hspace*{1.5em}``Reflections'': ``\textless how changes were applied and why\textgreater'',

\hspace*{1.5em}``Target Image Description'': ``\textless final\_target\_image\_description\textgreater''

\}

\textless /Response\textgreater

\medskip

\#\# Examples

Here are some examples for reference: ...

\end{tcolorbox}
\vspace{-5pt}
\caption{\textbf{Prompt of fine descriptions $P_\text{fine}$}}
\label{fig:prompt_fine}
\vspace{5pt}
\end{figure}
\begin{figure}
\centering
\vspace{-5pt}
\begin{tcolorbox}[title={\textbf{$D_{coarse}$}}]

\tiny
You are a visual captioning expert and editor. Using only the given ``Original Image'' and ``Modification Text,'' generate one Main Concept and one Target Caption.

\medskip

\#\# Definitions

Main Concept: A short, compressed, concise and visual phrase that summarizes the ``Original Image'' into its core, stable visual identity (visual anchor / core visual element) shared with the target after applying the ``Modification Text.''

\hspace{1.5em}- It must function as a compact summary of the reference image, so that someone can roughly imagine the image just from this phrase.

\hspace{1.5em}- Always include the main subject/object and, if necessary, one essential attribute or object that is structurally bound to it (e.g., ``woman with umbrella'', ``dog with ball'', ``singer in dress'').

\hspace{1.5em}- Do not include variable details such as color, number, background, style, pose, or temporary actions, unless the Modification Text explicitly marks them as preserved.

\hspace{1.5em}- If the ``Modification Text'' explicitly states that an attribute is preserved (e.g., same, still, remain, keep, unchanged), translate it into its concrete visual meaning (e.g., ``same style'' → ``greyscale photo'', ``same breed'' → ``bulldog'').

\hspace{1.5em}- Keep it short (3-6 words), minimal, and visual — enough to serve as an anchor for the Target Caption.

\medskip

Target Caption: A short, simple sentence that describes the target image, created by combining the Main Concept with the ``Modification Text.''

\hspace{1.5em}- You must use only the Main Concept and the Modification Text.
  
\hspace{1.5em}- Describe only what visually exists in the target image.
  
\hspace{1.5em}- For modifications that involve replacement, removal, or negation, include only the new or remaining elements, and completely exclude the removed ones.

\hspace{3em}- Do NOT include comparative or negation phrases like ``instead of,'' ``without,'' ``no.''

\hspace{3em}- Keep only the new or preserved element(s).

\hspace{1.5em}- Use the Main Concept as the anchor.

\hspace{1.5em}- Integrate only the visual modifications described in the ``Modification Text.''

\hspace{1.5em}- Do not add any extra verbs, actions, or contextual words.

\hspace{1.5em}- Keep it concise and visual.

\medskip

\#\# Critical Constraints (Visual-Only)

Include only elements that will visually appear in the image.
Do NOT include non-visual, abstract, or subjective terms (e.g., ``happy,'' ``moody,'' ``beautiful,'' ``cinematic'') unless they are concretely visualized and explicitly preserved.
Do NOT include intentions or instructions (e.g., ``to be changed,'' ``should become''); describe what is visually present in the target.

\medskip

\#\# Step 1: Main Concept

1) Read the ``Modification Text'' to determine what changes and thus what stays the same.

2) From the ``Original Image,'' extract the stable, shared visual subject/object (visual anchor) unaffected by the modification.

3) If the ``Modification Text'' explicitly marks an attribute as preserved (e.g., ``same style/background/pose''), include that attribute in concrete visual terms; otherwise exclude it.

4) Keep it minimal and intuitive so the concept stays valid both before and after the modification.

\medskip

\#\# Step 2: Target Caption

1) Start with the Main Concept.

2) Re-read the ``Modification Text'' and extract the visual modifications precisely:

\hspace{1.5em}- Addition: Add the new element(s) as described. (e.g., ``with a hat'' → ``man with a hat'')

\hspace{1.5em}- Replacement: Keep only the new element(s) and completely remove the replaced ones. (e.g., ``has plants instead of bird'' → ``plants'')

\hspace{1.5em}- Removal / Negation: Exclude the removed element(s) entirely. (e.g., ``without hat'' → ``''; ``no cars'' → ``'')

\hspace{1.5em}- Preservation: If explicitly stated (e.g., ``same, keep, remain, unchanged''), translate it into its concrete visual meaning and include it. (e.g., ``same style'' → ``grey photo'')

3) Combine the Main Concept with these modifications so that the final sentence contains only the visual elements that exist in the target image.

4) The final sentence must be short, concise, and purely visual — clear enough to be directly useful for image retrieval.

\medskip

\#\# Input Format

\textless Input\textgreater

\{

  ``Original Image'': \textless image\_text\_or\_url\textgreater ,
  
  ``Modification Text'': \textless modification\_text\textgreater
  
\}

\textless/Input\textgreater

\medskip

\#\# Strict Output Format
\textless Response\textgreater

\{

\hspace{1.5em}``Main Concept'': ``\textless main\_concept\textgreater '',
  
\hspace{1.5em}``Target Caption'': ``\textless target\_caption\textgreater''
  
\}

\textless/Response\textgreater

\medskip

\#\# Examples

Here are some examples for reference:
...

\end{tcolorbox}
\vspace{-5pt}
\caption{\textbf{Prompt of coarse descriptions $P_\text{coarse}$}}
\label{fig:prompt_coarse}
\vspace{-10pt}
\end{figure}

\newpage


\section{Additional Experiments}
\label{sec:additional_experiments}


\begin{table*}[!t]
\centering
\caption{\textbf{Extended ablation study across CLIP backbones.}
FD, CD, and Comp denote the fine-grained representation, coarse-grained representation, and composed-query anchor, respectively.
We report results on CIRCO, CIRR, and Fashion-IQ using ViT-B/32, ViT-L/14, and ViT-G/14.}
\label{table:supp_ablation}
\setlength{\tabcolsep}{4.0pt}
\resizebox{\textwidth}{!}{
\begin{tabular}{c ccc cc cc cc}
\Xhline{2\arrayrulewidth}
\noalign{\smallskip}

\multirow{2}{*}{\textbf{Backbone}} &
\multirow{2}{*}{\textbf{FD}} &
\multirow{2}{*}{\textbf{CD}} &
\multirow{2}{*}{\textbf{Comp}} &
\multicolumn{2}{c}{\textbf{CIRCO}} &
\multicolumn{2}{c}{\textbf{CIRR}} &
\multicolumn{2}{c}{\textbf{Fashion-IQ}} \\

\cmidrule(lr){5-6}
\cmidrule(lr){7-8}
\cmidrule(lr){9-10}

& & & &
\textbf{mAP@5} & \textbf{mAP@10} &
\textbf{R@5} & \textbf{R\textsubscript{sub}@1} &
\textbf{R@10} & \textbf{R@50} \\

\noalign{\smallskip}
\Xhline{2\arrayrulewidth}
\noalign{\smallskip}

\multirow{7}{*}{ViT-B/32}
& \cmark &        &        & 18.21 & 18.83 & 55.88 & 58.96 & 32.08 & 52.84 \\
&        & \cmark &        & 21.28 & 21.89 & 59.98 & 65.88 & 29.91 & 50.87 \\
&        &        & \cmark & 11.00 & 11.41 & 59.21 & 63.90 & 33.97 & 54.49 \\
& \cmark & \cmark &        & 22.95 & 23.76 & 62.29 & 65.16 & 35.57 & 57.02 \\
& \cmark &        & \cmark & 22.15 & 22.86 & 65.74 & 64.89 & 39.50 & 60.82 \\
&        & \cmark & \cmark & 23.53 & 24.05 & 64.46 & \textbf{68.07} & 37.75 & 59.03 \\
& \cmark & \cmark & \cmark & \textbf{25.07} & \textbf{25.81} & \textbf{67.28} & 67.66 & \textbf{40.65} & \textbf{62.01} \\

\noalign{\smallskip}
\hline
\noalign{\smallskip}

\multirow{7}{*}{ViT-L/14}
& \cmark &        &        & 21.95 & 22.59 & 58.17 & 60.19 & 32.38 & 52.42 \\
&        & \cmark &        & 27.28 & 28.11 & 62.78 & 68.07 & 29.10 & 49.83 \\
&        &        & \cmark & 13.62 & 14.15 & 62.19 & 66.55 & 36.66 & 56.17 \\
& \cmark & \cmark &        & 30.05 & 30.88 & 64.87 & 66.72 & 34.40 & 55.70 \\
& \cmark &        & \cmark & 27.59 & 28.47 & 67.59 & 67.35 & 39.90 & 60.79 \\
&        & \cmark & \cmark & 29.46 & 30.13 & 67.66 & \textbf{70.75} & 37.41 & 58.28 \\
& \cmark & \cmark & \cmark & \textbf{31.80} & \textbf{32.65} & \textbf{69.95} & 70.34 & \textbf{41.37} & \textbf{61.66} \\

\noalign{\smallskip}
\hline
\noalign{\smallskip}

\multirow{7}{*}{ViT-G/14}
& \cmark &        &        & 25.55 & 26.66 & 60.00 & 61.54 & 35.57 & 56.03 \\
&        & \cmark &        & 32.03 & 33.34 & 64.29 & 69.06 & 32.83 & 52.84 \\
&        &        & \cmark & 17.32 & 17.88 & 62.75 & 67.42 & 40.58 & 60.39 \\
& \cmark & \cmark &        & 34.77 & 35.95 & 67.04 & 68.15 & 38.43 & 59.35 \\
& \cmark &        & \cmark & 32.01 & 32.88 & 69.33 & 67.66 & 43.86 & 65.23 \\
&        & \cmark & \cmark & 34.31 & 34.99 & 69.74 & \textbf{70.58} & 41.49 & 63.22 \\
& \cmark & \cmark & \cmark & \textbf{37.47} & \textbf{38.63} & \textbf{71.40} & 70.27 & \textbf{45.68} & \textbf{65.74} \\

\noalign{\smallskip}
\Xhline{2\arrayrulewidth}
\end{tabular}
}
\end{table*}
\subsection{Ablation Studies}
Table~\ref{table:supp_ablation} provides extended ablation results across different CLIP backbones.
The results show a consistent trend across ViT-B/32, ViT-L/14, and ViT-G/14.
Using all three components generally achieves the best performance across CIRCO, CIRR, and Fashion-IQ, confirming the effectiveness of jointly exploiting fine-grained descriptions, coarse-grained descriptions, and the composed-query anchor.
The single-component results indicate that each component captures different retrieval cues.
CD is particularly effective on CIRCO and CIRR, suggesting that coarse semantic abstraction is useful for open-domain composed retrieval.
In contrast, Comp shows strong performance on Fashion-IQ, indicating that preserving the original reference-modification information is important for fine-grained fashion retrieval.
Pairwise combinations further improve performance over most single-component settings, demonstrating the complementarity among the three components.
Overall, the extended ablation results confirm that the performance gains of DCR-TSF are stable across backbone scales.

\subsection{Analysis of Fusion Strategies}
To examine whether the effectiveness of TSF depends on the choice of fusion operator, we compare normalized weighted averaging, sequential Slerp, and the weighted Karcher mean.
For a controlled comparison, all methods fuse the same three representations, $T_{\text{fine}}$, $T_{\text{coarse}}$, and $Q_{\text{comp}}$, using identical fusion weights.
As shown in Table~\ref{table:supp_fusion}, all three fusion strategies achieve nearly identical performance across CIRCO, CIRR, and Fashion-IQ.
This indicates that the performance gains of TSF primarily arise from integrating the complementary information provided by the three representations, rather than from a particular fusion operator.
We adopt the weighted Karcher mean as a principled intrinsic formulation for multi-way fusion on the unit hypersphere.
\begin{table*}[!t]
\centering
\caption{\textbf{Comparison of fusion strategies across CLIP backbones.}
Results are reported on CIRCO, CIRR, and Fashion-IQ with ViT-B/32 and ViT-L/14.}
\label{table:supp_fusion}
\setlength{\tabcolsep}{5.0pt}
\resizebox{0.78\textwidth}{!}{
\begin{tabular}{c c c c c}
\Xhline{2\arrayrulewidth}
\noalign{\smallskip}

\multirow{2}{*}{\textbf{Backbone}} &
\multirow{2}{*}{\textbf{Fusion Strategy}} &
\textbf{CIRCO} &
\textbf{CIRR} &
\textbf{Fashion-IQ} \\

\cmidrule(lr){3-3}
\cmidrule(lr){4-4}
\cmidrule(lr){5-5}

& &
mAP@5 &
R@5 &
R\textsubscript{mean} \\

\noalign{\smallskip}
\Xhline{2\arrayrulewidth}
\noalign{\smallskip}

\multirow{3}{*}{ViT-B/32}
& Weighted Avg.     & \textbf{25.11} & 67.18 & 51.32 \\
& Sequential Slerp  & 25.05 & 67.22 & 51.31 \\
& Karcher Mean      & 25.07 & \textbf{67.28} & \textbf{51.33} \\

\noalign{\smallskip}
\hline
\noalign{\smallskip}

\multirow{3}{*}{ViT-L/14}
& Weighted Avg.     & \textbf{31.92} & \textbf{69.95} & 51.46 \\
& Sequential Slerp  & 31.56 & 69.88 & 51.41 \\
& Karcher Mean      & 31.80 & \textbf{69.95} & \textbf{51.52} \\

\noalign{\smallskip}
\Xhline{2\arrayrulewidth}
\end{tabular}
}
\end{table*}

\subsection{Effect of Reasoning Backbone}
To examine whether the performance improvement stems from the use of a stronger reasoning backbone, Table~\ref{table:supp_mllm} summarizes the reasoning backbones used by existing training-free ZS-CIR methods.
In particular, when using the same Qwen2-VL-72B backbone as CoTMR, DCR alone, without TSF, consistently outperforms CoTMR.
These results demonstrate that the performance improvement is not attributable to the use of a stronger reasoning backbone.
\begin{table}[!t]
\centering
\caption{\textbf{Comparison of reasoning backbones on the CIRCO test set with CLIP ViT-G/14.}}
\label{table:supp_mllm}
\setlength{\tabcolsep}{5.0pt}
\resizebox{0.82\linewidth}{!}{
\begin{tabular}{l l cc}
\Xhline{2\arrayrulewidth}
\noalign{\smallskip}

\textbf{Method} &
\textbf{Reasoning Backbone} &
\textbf{mAP@5} &
\textbf{mAP@10} \\

\noalign{\smallskip}
\Xhline{2\arrayrulewidth}
\noalign{\smallskip}

CIReVL~\cite{CIReVL} &
BLIP-2~\cite{BLIP2} + GPT-3.5-Turbo~\cite{gpt-3.5} &
26.77 & 27.59 \\

OSrCIR~\cite{OSrCIR} &
GPT-4o~\cite{gpt-4} &
30.47 & 31.14 \\

SEIZE~\cite{SEIZE} &
BLIP-2~\cite{BLIP2} + GPT-3.5-Turbo~\cite{gpt-3.5} &
32.46 & 33.77 \\

CoTMR~\cite{CoTMR} &
Qwen2-VL-72B~\cite{Qwen2} &
32.23 & 32.72 \\

\noalign{\smallskip}
\hline
\noalign{\smallskip}

DCR (Ours) &
Qwen2-VL-72B~\cite{Qwen2} &
34.38 & \textbf{36.11} \\

\baseline{DCR (Ours)} &
\baseline{Qwen2.5-VL-32B~\cite{Qwen2.5}} &
\baseline{\textbf{34.77}} &
\baseline{35.95} \\

\noalign{\smallskip}
\Xhline{2\arrayrulewidth}
\end{tabular}
}
\end{table}

\subsection{Hyperparameter Sensitivity Analysis}
Table~\ref{table:supp_weight} evaluates the sensitivity of DCR-TSF to the fusion weights by assigning equal weights of $1/3$ to all three representations.
DCR-TSF maintains strong performance across CIRCO, CIRR, and Fashion-IQ even without dataset-specific weight configurations.
These results show that the performance of DCR-TSF does not rely on carefully tuned fusion weights, demonstrating its robustness to the choice of fusion weights.
\begin{table*}[!t]
\centering
\caption{\textbf{Sensitivity analysis of fusion weights.}
Comparison of equal and dataset-specific fusion weights on CIRCO, CIRR, and Fashion-IQ using CLIP ViT-B/32 and ViT-L/14.}
\label{table:supp_weight}
\setlength{\tabcolsep}{5.0pt}
\resizebox{0.78\textwidth}{!}{
\begin{tabular}{c c c c c}
\Xhline{2\arrayrulewidth}
\noalign{\smallskip}

\multirow{2}{*}{\textbf{Backbone}} &
\multirow{2}{*}{\textbf{Fusion Weights}} &
\textbf{CIRCO} &
\textbf{CIRR} &
\textbf{Fashion-IQ} \\

\cmidrule(lr){3-3}
\cmidrule(lr){4-4}
\cmidrule(lr){5-5}

& &
mAP@5 &
R@5 &
R\textsubscript{mean}\\

\noalign{\smallskip}
\Xhline{2\arrayrulewidth}
\noalign{\smallskip}

\multirow{2}{*}{ViT-B/32}
& Equal Weights    & 25.02 & 66.77 & 50.81 \\
& Dataset-specific & \textbf{25.07} & \textbf{67.28} & \textbf{51.33} \\

\noalign{\smallskip}
\hline
\noalign{\smallskip}

\multirow{2}{*}{ViT-L/14}
& Equal Weights    & 31.60 & 69.25 & 50.46 \\
& Dataset-specific & \textbf{31.80} & \textbf{69.95} & \textbf{51.52} \\

\noalign{\smallskip}
\Xhline{2\arrayrulewidth}
\end{tabular}
}
\end{table*}

\subsection{Analysis of Modality Gap}
Recent studies have identified a modality gap in multimodal embedding models such as CLIP, where visual and textual representations can remain separated despite being mapped into a shared embedding space~\cite{modalitygap, 3_hierspeech, 2_hierspeech++}.
We further analyze how reducing this modality gap affects the retrieval performance of DCR-TSF.
To this end, we employ I0T\textsubscript{post}~\cite{I0T}, a training-free method that reduces the modality gap through post-hoc alignment.
As a simple diagnostic, we measure the modality gap as the $\ell_2$ distance between the mean visual and textual embeddings.
As shown in Table~\ref{table:supp_modality}, I0T\textsubscript{post} substantially reduces the modality gap, while retrieval performance does not consistently improve across datasets.
These results indicate that reducing the modality gap alone does not necessarily lead to improved composed image retrieval performance.
\begin{table}[!t]
\centering
\caption{\textbf{Analysis of modality gap reduction.}
Comparison of retrieval performance and modality gap with and without I0T\textsubscript{post} on the validation splits of three benchmarks.
Values are reported as DCR-TSF / I0T\textsubscript{post}.}
\label{table:supp_modality}
\setlength{\tabcolsep}{5.0pt}
\resizebox{0.78\linewidth}{!}{
\begin{tabular}{l c c c}
\Xhline{2\arrayrulewidth}
\noalign{\smallskip}

&
\textbf{CIRCO (mAP@5)} &
\textbf{CIRR (R@5)} &
\textbf{Fashion-IQ (R\textsubscript{mean})} \\

\noalign{\smallskip}
\Xhline{2\arrayrulewidth}
\noalign{\smallskip}

Performance $\uparrow$
& \textbf{35.35} / 34.63
& \textbf{72.61} / 70.80
& 55.71 / \textbf{58.30} \\

Modality Gap $\downarrow$
& 0.5524 / \textbf{0.0084}
& 0.5366 / \textbf{0.0133}
& 0.5990 / \textbf{0.0195} \\

\noalign{\smallskip}
\Xhline{2\arrayrulewidth}
\end{tabular}
}
\end{table}

\subsection{Comparative Analysis of Caption Ensemble and DCR}
\begin{figure}[t]
    \centering

    \begin{minipage}[t]{0.44\linewidth}
        \centering
        \includegraphics[
            width=\linewidth,
            height=3.0cm
        ]{figs/mAP10_caption_compare_1.png}
        \vspace{-3mm}

        {\small (a) CIRCO}
    \end{minipage}
    \hspace{0.03\linewidth}
    \begin{minipage}[t]{0.44\linewidth}
        \centering
        \includegraphics[
            width=\linewidth,
            height=3.0cm
        ]{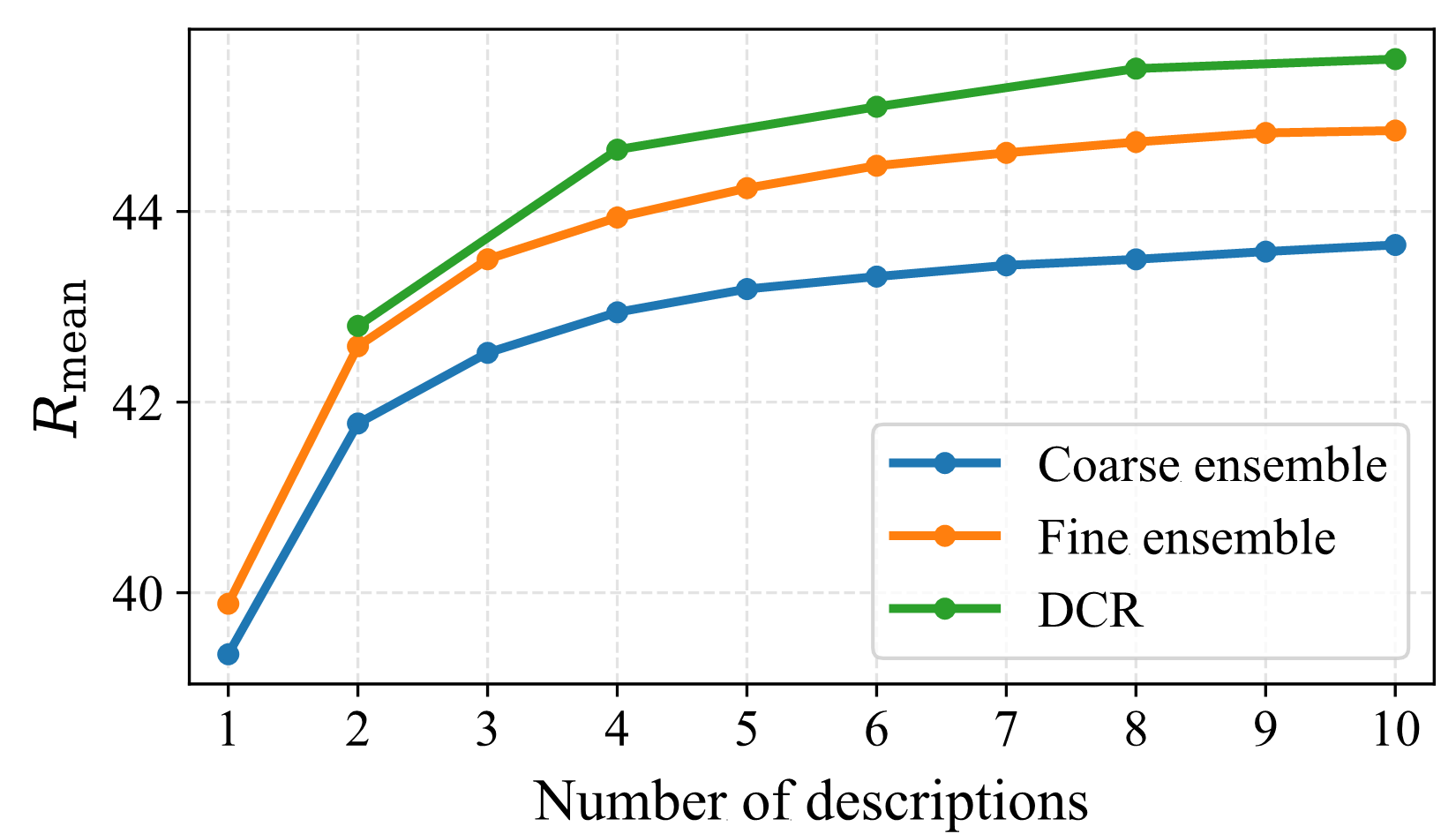}
        \vspace{-3mm}

        {\small (b) Fashion-IQ}
    \end{minipage}

    \caption{\textbf{Comparison of DCR with caption ensemble baselines on CIRCO and Fashion-IQ.}}
    \label{fig:caption_ensemble_supp}
\end{figure}

We further investigate whether the performance gains of DCR can be explained by simply ensembling multiple captions.
To this end, we compare DCR with caption ensemble baselines constructed from descriptions at the same granularity.
As shown in Fig.~\ref{fig:caption_ensemble_supp}, the relative effectiveness of coarse- and fine-grained caption ensembles varies across datasets, indicating that the preferred level of description granularity can depend on the characteristics of each benchmark.
Despite this variation, DCR consistently outperforms the corresponding caption ensemble baselines on both CIRCO and Fashion-IQ.
These results suggest that the benefit of DCR does not arise merely from aggregating multiple captions, but from effectively leveraging complementary information across descriptions with different levels of granularity.

\section{Additional Qualitative Results}
\label{sec:additional_qualitative_results}

\subsection{Retrieval Results}
\label{subsec:additional_retrieval_results}

To further illustrate the behavior of our method beyond quantitative metrics, we present qualitative retrieval results on Fashion-IQ and CIRR.
Each example compares four retrieval settings: \emph{Fine-Grained Description}, \emph{Coarse-Grained Description}, \emph{Composed-query Anchor}, and \emph{\modelname}.
The ground-truth image is highlighted in green.

\begin{figure}[!ht]
    \centering
    \begin{subfigure}[t]{\linewidth}
        \centering
        \includegraphics[width=\linewidth]{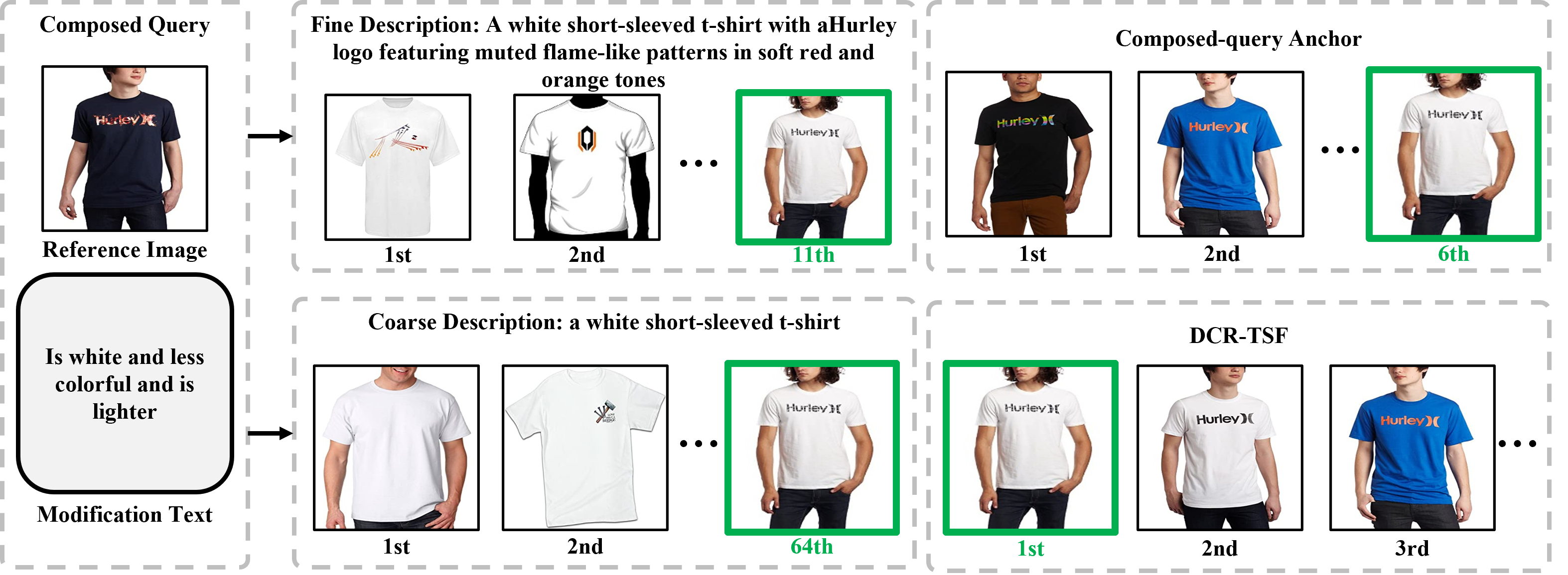}
        \caption{Shirt}
    \end{subfigure}%
    \\
    \begin{subfigure}[t]{\linewidth}
        \centering
        \includegraphics[width=\linewidth]{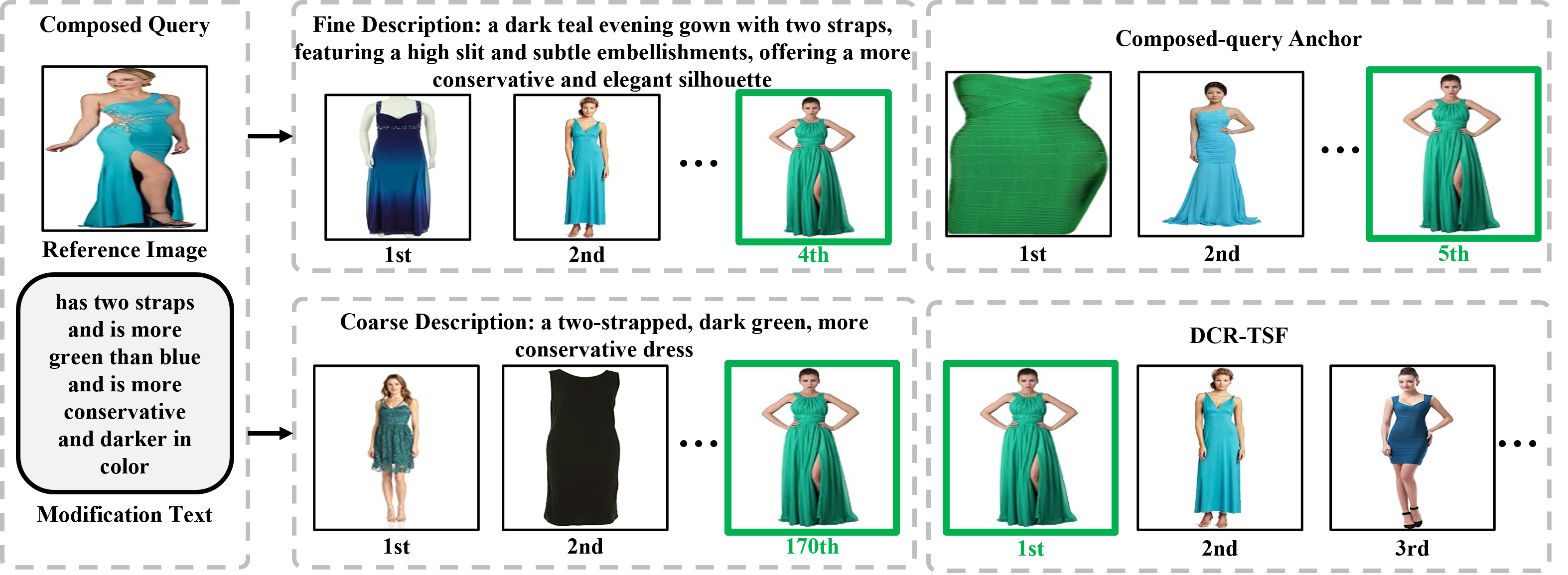}
        \caption{Dress}
    \end{subfigure}%
    \\
    \begin{subfigure}[t]{\linewidth}
        \centering
        \includegraphics[width=\linewidth]{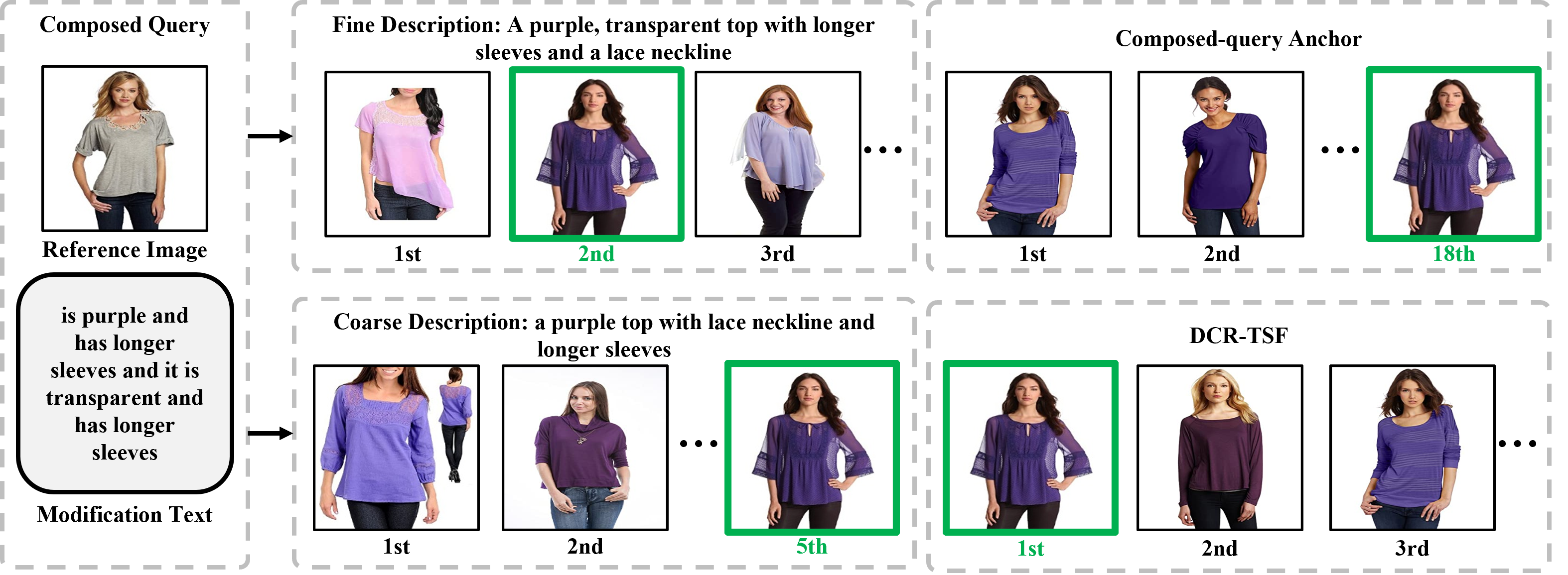}
        \caption{Tops\&Tee}
    \end{subfigure}%

    \caption{\textbf{Qualitative comparison on Fashion-IQ with ViT-G/14.}
    Each example shows a reference image, its modification text, and the retrieved images with their ranks under four settings:
    \emph{Fine-Grained Description}, \emph{Coarse-Grained Description}, \emph{Composed-query Anchor}, and \emph{\modelname}.
    The ground-truth image is highlighted in green.}
    \label{fig:ret_results_fiq}
\end{figure}

\noindent \textbf{Fashion-IQ.}
Fig.~\ref{fig:ret_results_fiq} shows representative examples from the Shirt, Dress, and Tops\&Tee categories.
Fashion-IQ requires fine-grained attribute understanding, since the modification text often describes detailed changes in color, shape, texture, sleeve length, or garment style.
Fine-grained descriptions and the composed-query anchor help preserve such detailed reference-modification cues, while coarse-grained descriptions provide complementary garment-level semantics.
By combining these cues, \modelname\ more effectively captures both detailed visual attributes and the intended modification, ranking the ground-truth item higher.

\begin{figure}[!t]
    \centering
    \includegraphics[width=\linewidth]{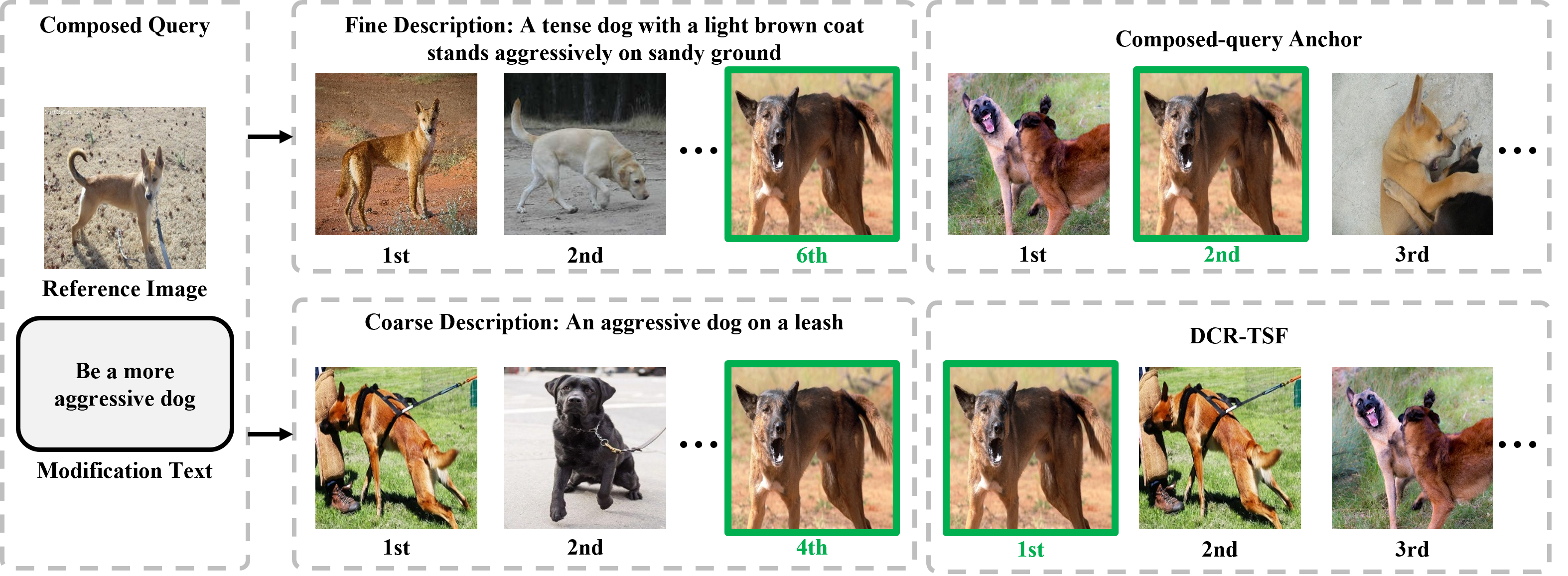}
    \caption{\textbf{Qualitative comparison on CIRR with ViT-G/14.}
    Each example shows a reference image, its modification text, and the retrieved images with their ranks under four settings:
    \emph{Fine-Grained Description}, \emph{Coarse-Grained Description}, \emph{Composed-query Anchor}, and \emph{\modelname}.
    The ground-truth image is highlighted in green.}
    \label{fig:ret_results_cirr}

    \vspace{20pt}

    \begin{subfigure}[t]{0.49\linewidth}
        \centering
        \includegraphics[width=\linewidth]{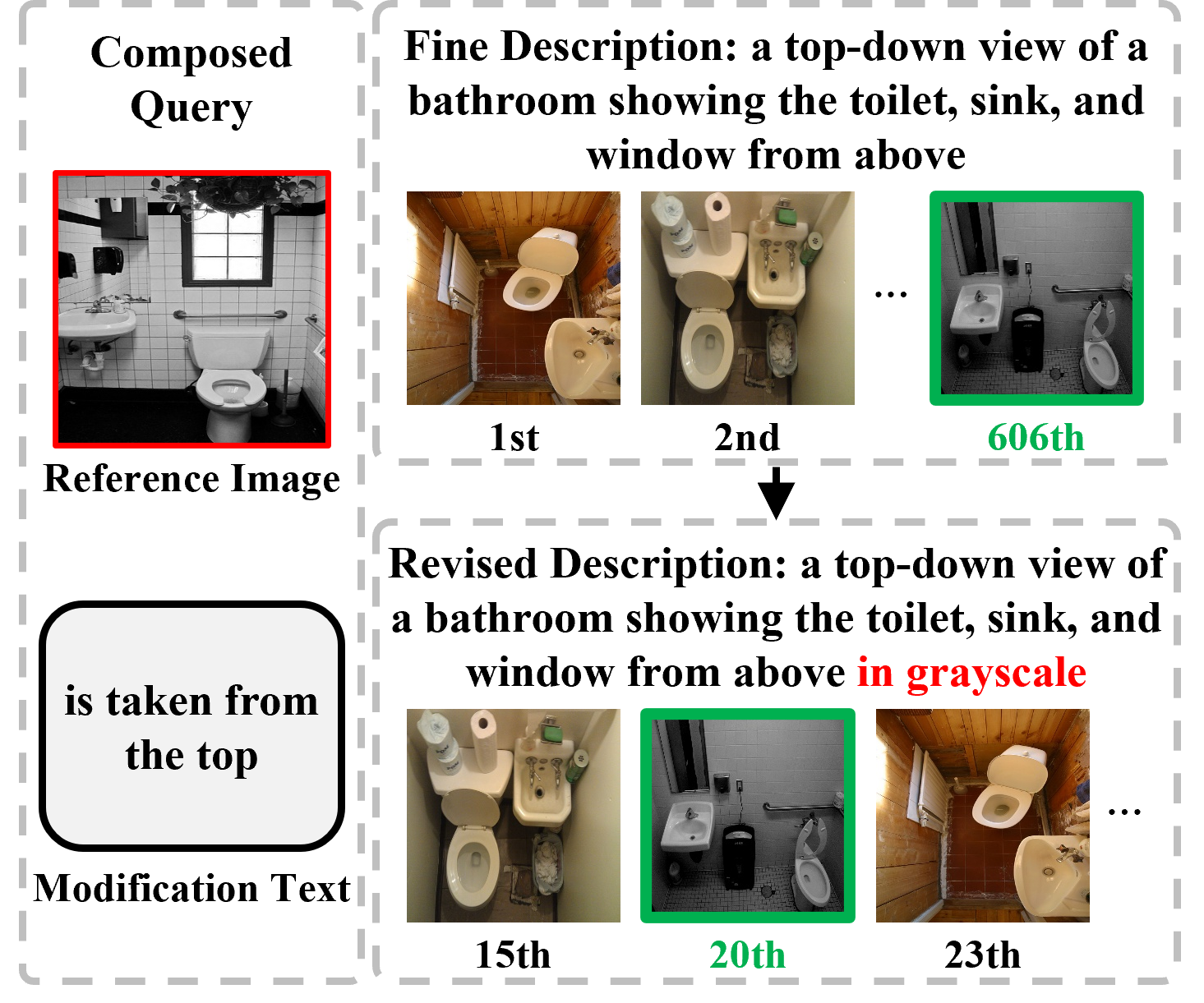}
    \end{subfigure}
    \hfill
    \begin{subfigure}[t]{0.49\linewidth}
        \centering
        \includegraphics[width=\linewidth]{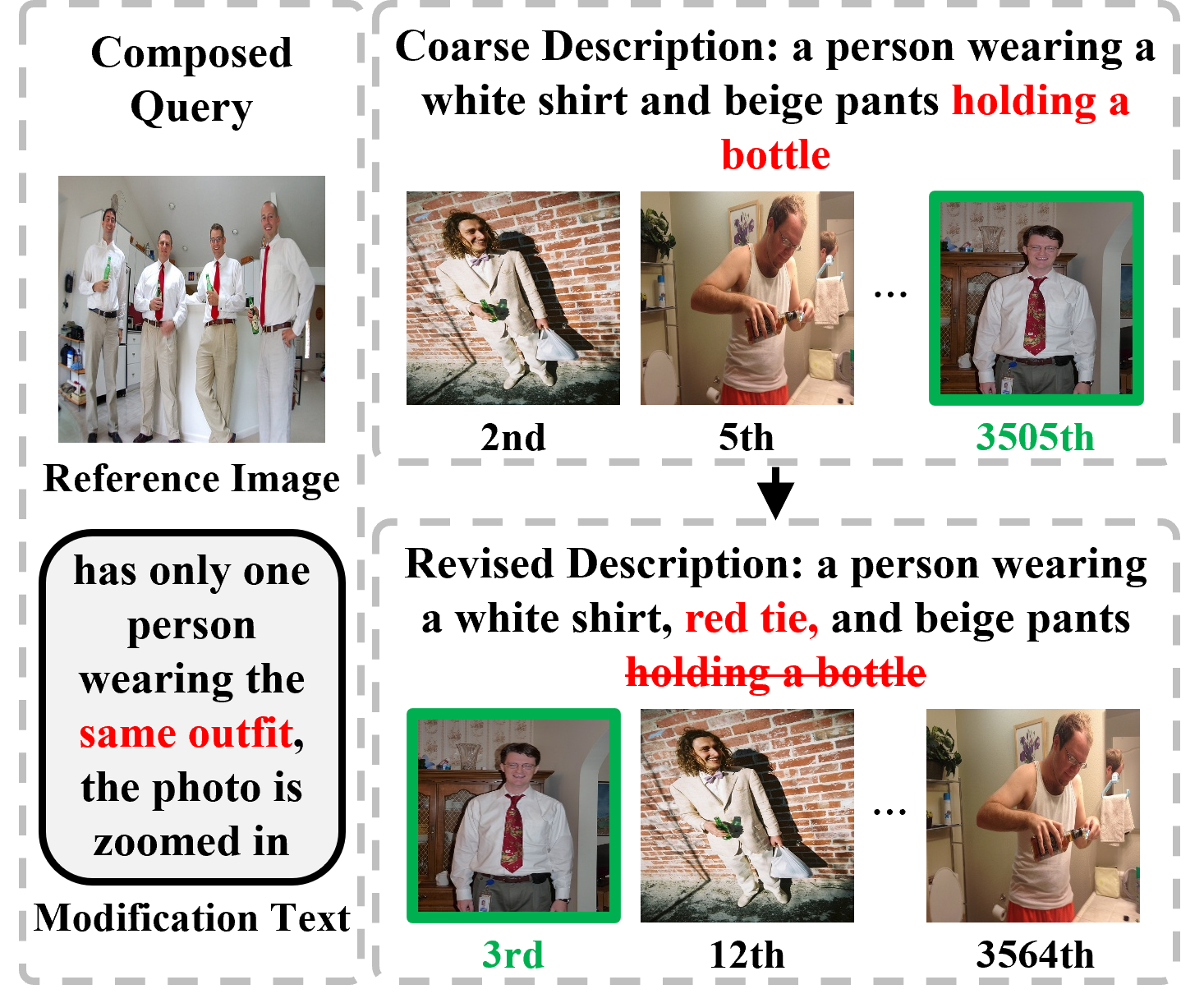}
    \end{subfigure}
    \caption{\textbf{Failure cases caused by MLLM generation errors on CIRCO with ViT-G/14.}
    These examples illustrate failure cases caused by hallucinated or inaccurate MLLM-generated descriptions.
    When incorrect semantics are introduced into the generated target description, they can propagate to the retrieval stage and degrade the target ranking.
    After manually correcting the erroneous attributes, the ranking improves substantially, highlighting the importance of description quality.}
    \label{fig:failure_case_mllm}
\end{figure}

\noindent \textbf{CIRR.}
Fig.~\ref{fig:ret_results_cirr} shows qualitative examples from CIRR.
CIRR contains diverse natural images, where the modification text often requires changes in object category, action, or scene context.
Fine- and coarse-grained descriptions capture complementary target semantics, while the composed-query anchor preserves the original reference-modification condition.
By combining these cues, \modelname\ improves the rank of the ground-truth image.


\begin{figure}[!t]
    \centering

    \begin{subfigure}[t]{0.49\linewidth}
        \centering
        \includegraphics[width=\linewidth]{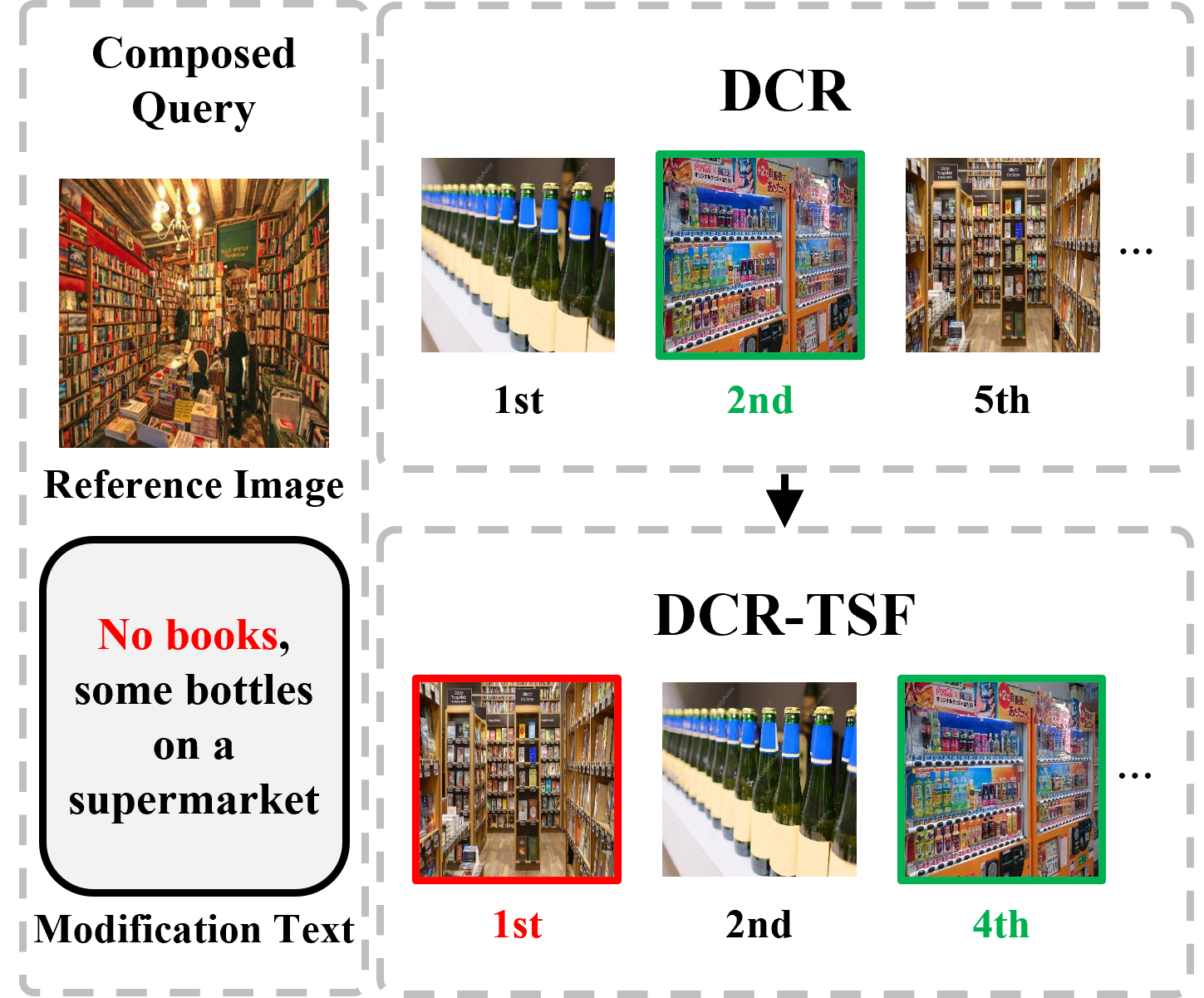}
    \end{subfigure}
    \hfill
    \begin{subfigure}[t]{0.49\linewidth}
        \centering
        \includegraphics[width=\linewidth]{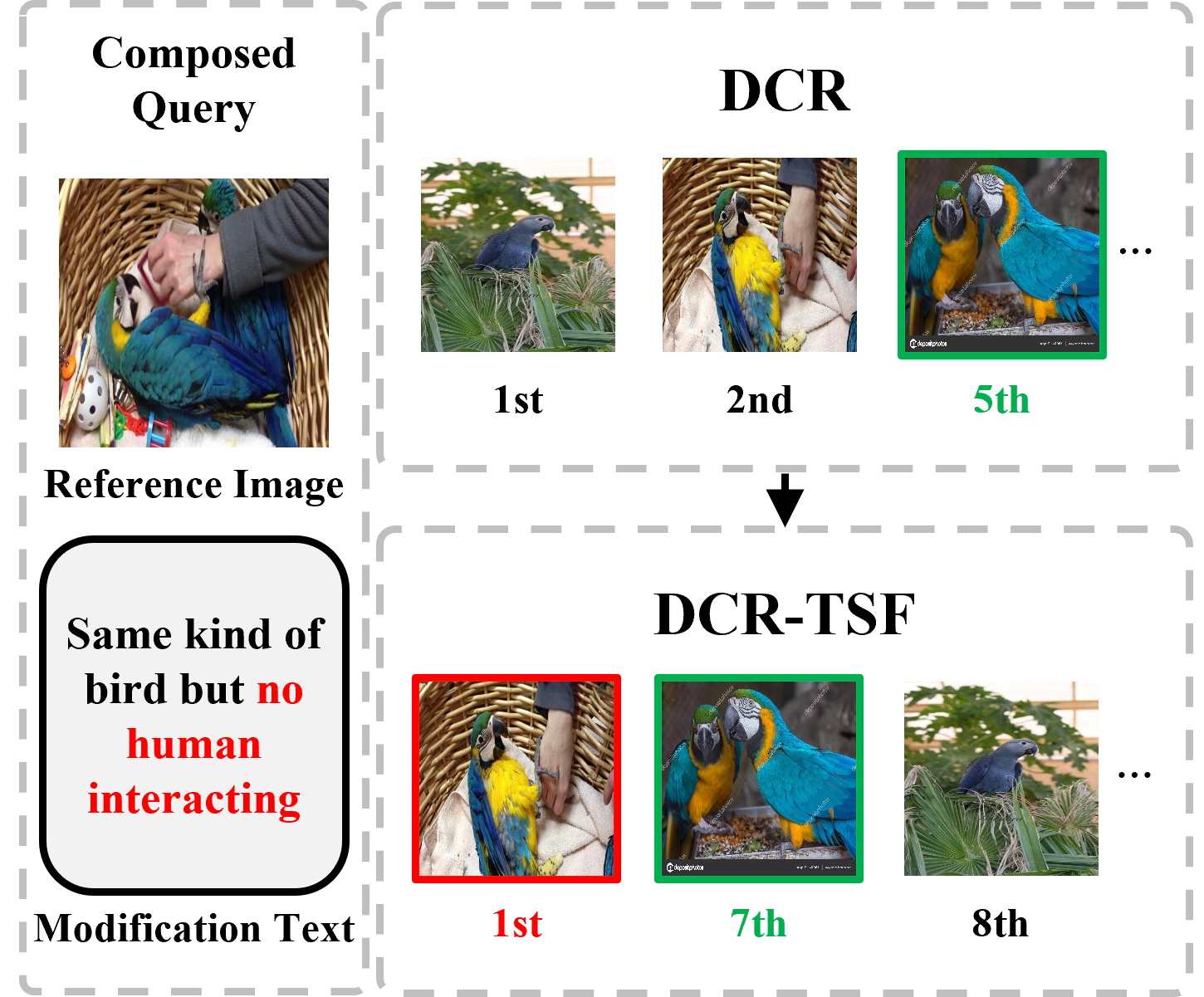}
    \end{subfigure}

    \caption{\textbf{Failure cases caused by negation on CIRR with ViT-G/14.}
    CLIP assigns high similarity to images containing objects or interactions explicitly excluded by negated modification texts (e.g., ``no books'' or ``no human interacting'').
    Since TSF directly incorporates the modification text, this limitation propagates to retrieval and can rank such images above the ground-truth target.}
    \label{fig:failure_case_negation}
\end{figure}

\subsection{Failure Case Analysis}
Despite the strong performance, our method still inherits limitations from both the MLLM and CLIP used in training-free ZS-CIR.
Figs.~\ref{fig:failure_case_mllm} and~\ref{fig:failure_case_negation} illustrate representative failure cases.

\noindent \textbf{MLLM generation errors.}
Similar to existing MLLM-based ZS-CIR methods, our method relies on generated target descriptions.
Therefore, when the MLLM produces incorrect or hallucinated fine- or coarse-grained descriptions, these errors can be directly propagated to the retrieval stage.
As shown in Fig.~\ref{fig:failure_case_mllm}, correcting the generated descriptions leads to clear rank improvements, indicating that description quality remains an important bottleneck.

\noindent \textbf{Failure on negation.}
Fig.~\ref{fig:failure_case_negation} shows failure cases arising from CLIP's difficulty in handling negation in the original modification text.
For example, expressions such as ``no books'' or ``no human interacting'' may not be sufficiently reflected in the CLIP similarity score.
Since TSF directly incorporates the original modification text, this limitation can propagate to the retrieval stage, causing images that violate the negated condition to be ranked above the ground-truth target.

\end{document}